\documentclass[nohyperref]{article}

\usepackage{microtype}
\usepackage{graphicx}
\usepackage{subfigure}
\usepackage{multirow}
\usepackage{booktabs} 
\usepackage{caption}
\usepackage{algorithm}
\usepackage{listings}
\usepackage[table]{xcolor}

\usepackage{hyperref}

\usepackage[accepted]{icml2022}

\usepackage{amsmath}
\usepackage{amssymb}
\usepackage{mathtools}
\usepackage{amsthm}

\usepackage[capitalize,noabbrev]{cleveref}

\theoremstyle{plain}

\theoremstyle{definition}

\theoremstyle{remark}

\usepackage[textsize=tiny]{todonotes}

\icmltitlerunning{{Position Prediction as an Effective Pretraining Strategy}}

\begin{document}

\twocolumn[
\icmltitle{Position Prediction as an Effective Pretraining Strategy}



\icmlsetsymbol{equal}{*}

\begin{icmlauthorlist}
\icmlauthor{Shuangfei Zhai}{apple}
\icmlauthor{Navdeep Jaitly}{apple}
\icmlauthor{Jason Ramapuram}{apple}
\icmlauthor{Dan Busbridge}{apple}
\icmlauthor{Tatiana Likhomanenko}{apple}
\icmlauthor{Joseph Yitan Cheng}{apple}
\icmlauthor{Walter Talbott}{apple}
\icmlauthor{Chen Huang}{apple}
\icmlauthor{Hanlin Goh}{apple}
\icmlauthor{Joshua Susskind}{apple}
\end{icmlauthorlist}

\icmlaffiliation{apple}{Apple Inc}

\icmlcorrespondingauthor{Shuangfei Zhai}{szhai@apple.com}

\icmlkeywords{Machine Learning, ICML}

\vskip 0.3in
]



\printAffiliationsAndNotice{}  

\begin{abstract}
Transformers \cite{transformer} have gained increasing popularity in a wide range of applications, including Natural Language Processing (NLP), Computer Vision and Speech Recognition, because of their powerful representational capacity. However, harnessing this representational capacity effectively requires a large amount of data, strong regularization, or both, to mitigate overfitting. Recently, the power of the Transformer has been unlocked by self-supervised pretraining strategies based on masked autoencoders
which rely on reconstructing masked inputs, directly, or contrastively from unmasked content. This pretraining strategy which has been used in BERT models in NLP \cite{bert}, Wav2Vec models in Speech \cite{wv2v2} and, recently, in MAE models in Vision \cite{beit, mae}, forces the model to learn about relationships between the content in different parts of the input using autoencoding related objectives.  In this paper, we propose a novel, but surprisingly simple alternative to content reconstruction~-- that of predicting locations from content, without providing positional information for it. Doing so requires the Transformer to understand the positional relationships between different parts of the input, from their content alone. This amounts to an efficient implementation where the pretext task is a classification problem among all possible positions for each input token. We experiment on both Vision and Speech benchmarks, where our approach brings improvements over strong supervised training baselines and is comparable to modern unsupervised/self-supervised pretraining methods. Our method also enables Transformers trained without position embeddings to outperform ones trained with full position information.       
 

\end{abstract}

\section{Introduction}
Transformers \cite{transformer} have become a unified architecture in NLP, Computer Vision and Speech. 
Their high capacity and lack of domain specific inductive biases means Transformers require large amounts of training data to achieve good generalization. 
One effective remedy, first developed in the NLP community, is unsupervised pretraining. 
For example, BERT \cite{bert} trains a Transformer with unlabeled text by solving masked token prediction. 
This greatly benefits downstream applications, and has become the standard approach for various NLP tasks. 

Recently, there have been a few attempts to apply the BERT pretraining idea to Computer Vision, with Vision Transformers (ViTs) \cite{vit} being the backbone architecture. In particular, BEiT~\cite{beit} converts image patches to discrete tokens with a separately trained VQVAE \cite{vqvae}. This makes it possible to use the same cross entropy loss for masked image patch prediction as for token prediction in BERT. MAE \cite{mae} further simplifies the recipe of BEiT by directly predicting the masked patches with a regression loss in the pixel space. 

\begin{figure*}
    \centering
    \includegraphics[scale=0.2]{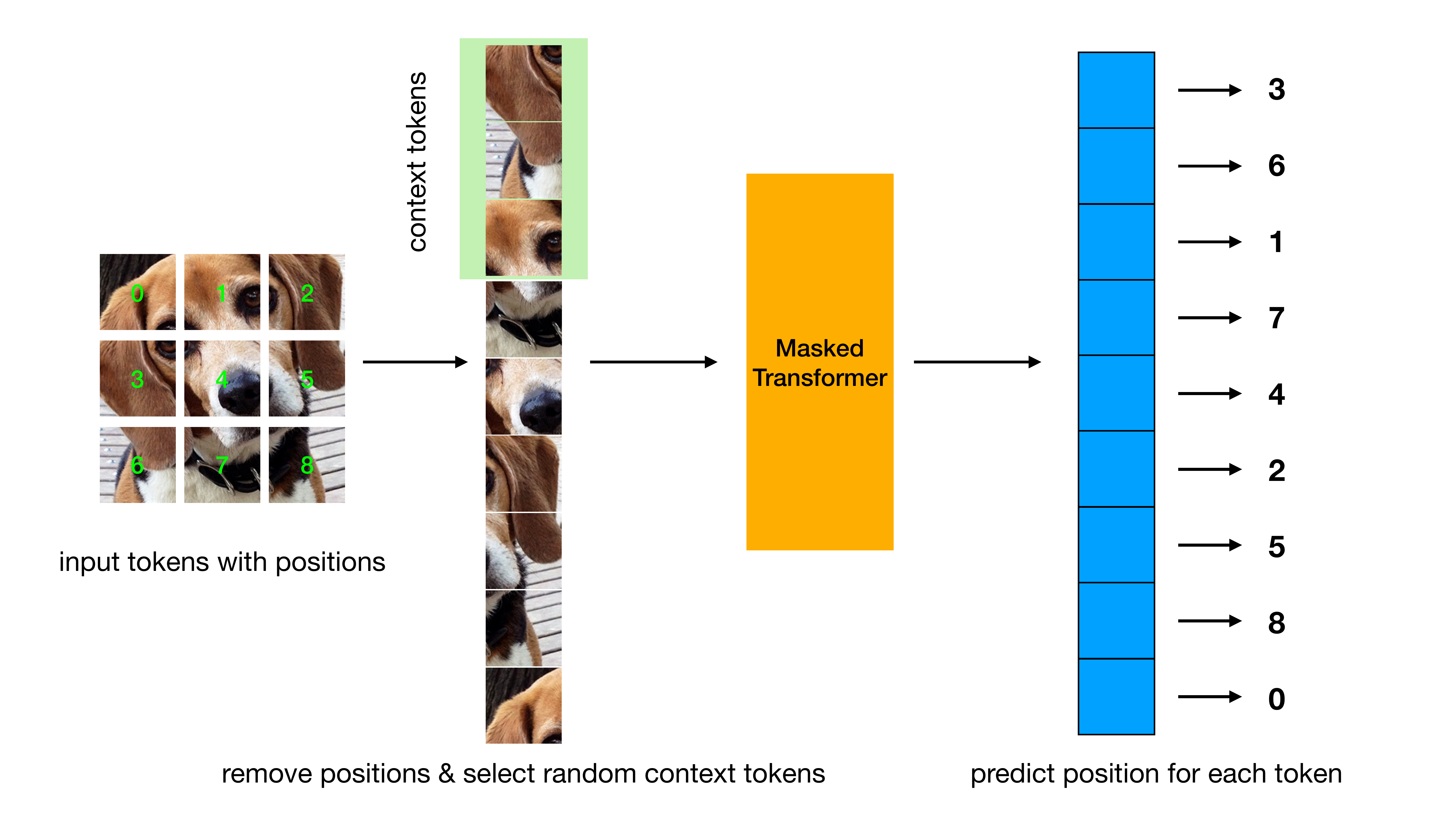}
    \caption{Illustration of our method MP3 on images. MP3 removes the position information for all tokens (image patches); it then randomly select a subset of tokens as context tokens. A \textbf{Masked Transformer} is used, where in each attention layer only context tokens contribute to the keys and values, and all tokens contribute to the queries. Each token predicts its position with a linear classifier head.}
    \label{fig:model}
\end{figure*}

In this paper, we propose a simple and effective approach for Transformer pretraining that removes the need for reconstructing dense patch values. Our idea is inspired by the observation that Transformers are relatively insensitive to the order of input tokens. In \cite{naseer2021intriguing}, it is shown that pretrained ViTs demonstrate strong robustness to image patch shuffling perturbation at test time. \cite{wordorder} shows that training the BERT model with randomly shuffled word order gives surprisingly competitive performance. \cite{mocov3} also suggests that a ViT without positional embeddings shows only a small degradation in the linear probing task for self-supervised learning. This evidence suggests that much of the power of Transformers results from the ability to reason about the co-occurrence of the set of unordered input tokens. We thus ask the question: \emph{How much can unsupervised pretraining learn using only contents for prediction?} This motivates us to formulate a novel pretraining strategy, explained as follows. 

In the pretraining phase, the model (e.g., a ViT) receives a set of tokens (e.g., image patches) but not their positions, and the pretext task is to recover the position of each input token, cast as a classification problem among all positions. By doing so, we formulate a special case of masked autoencoder, where the positions, rather than tokens, are removed from the input and predicted by the model. This training objective can also be interpreted as training a Set Transformer \cite{set_transformer} to solve a Jigsaw puzzle \cite{jigsaw}. In order to solve the task, the Transformer needs to reason about the high order interaction of the input tokens, which amounts to understanding the underlying semantics (e.g., part-whole relationship of the given object) represented by the inputs. Empirically, we have found that large Transformers can often achieve near perfect accuracy on the position prediction task\footnote{Except for Speech, where a patch (audio frame) is a small part of the full sequence, and it is much more challenging without providing some reference points with known positions.}. We then propose to increase the task's difficulty by selecting a random subset of tokens as context, and modify the attention layers such that only context tokens are used as keys and values. In this way, the Transformer not only needs to order the context tokens, but also infer the positions for masked out tokens by querying into the context. We hence dub our method \textbf{MP3}, denoting Masked Patch Position Prediction.

During finetuning, we disable token masking and add absolute positional embeddings in the same way as standard Transformers. We then remove the linear position prediction head and replace it with a linear head for the downstream task (e.g., classification). All the parameters are updated for a desired number of finetuning steps with the downstream task's training objective.

MP3 amounts to a simple implementation. In the pretraining phase, no additional modules are needed other than a linear head with $d \times n$ parameters, where $d$ is the model's feature dimension and $n$ is the number of positions. The training objective is simply the cross entropy loss. Also, thanks to the context masking, full self-attention is reduced to sparse attention, which effectively makes the pretraining cost lower than that of finetuning.

We conduct experiments on both Vision and Speech tasks. MP3 consistently improves the performance of Transformer models compared to strong supervised training baselines, and matches other more sophisticated unsupervised/self-supervised pretraining methods, despite is simplicity. Remarkably, MP3 enables strong finetuning performance even without using position embeddings, sometimes outperforming the supervised training baselines by a large margin. 


\section{Related Work}
\textbf{Denoising Autoencoders} (DAEs). DAEs \cite{vincent2010stacked} are well studied models in the context of unsupervised pretraining. The idea is to reconstruct the inputs given noisy versions of themselves. Masked autoencoder (MAE) is a special case of DAE, where part of the inputs are masked out (with a multiplicative Bernoulli noise). When combined with Transformers, MAEs have shown great success as an unsupervised pretraining technique, with BERT \cite{bert}, BEiT \cite{beit} and MAE \cite{mae} as notable examples. MP3 can also be viewed as a special case of MAEs, but it masks out the positional information (and optionally input tokens), rather than input tokens. The reconstruction objective is then turned into a sorting task, which has very different implications than reconstructing missing tokens given positions. 

\textbf{Self-supervised learning with order prediction}. Unsupervised feature learning with order prediction of image patches is first proposed in \cite{jigsaw}, and then extended in followup works such as \cite{lee2017unsupervised,ahsan2019video,xu2019self,santa2018visual,el2019skip}. What these works share is that they often adopt a CNN based encoder for an image patch or a video clip, and an MLP based prediction network to output the correct order of a set of inputs (except \cite{el2019skip} which uses order prediction to approximate future prediction in videos). The output of these methods is then a \emph{local} representation for image patches or video clips, as the order prediction network is discarded. This is in stark contrast with our work, MP3 focuses on learning the \emph{global} representation via attention. This is only made possible by the powerful Transformer architecture, which focuses on learning the interactions between input elements, and the same global knowledge is transferred to downstream tasks in the finetuning step.

\textbf{Importance of positional embedding in Transformers}. Positional embeddings (PEs) are of unique importance to Transformers, and improving PEs is an active research area, see \cite{dufter2021position} for an overview. However, it is empirically observed that the performance of Tranformers is surprisingly robust to the order of the inputs. For ViTs, \cite{naseer2021intriguing} shows that pretrained ViTs suffer much less from patch shuffling perturbations than CNNs. \cite{wordorder} shows that masked language models perform well even when trained with shuffled sentences. \cite{mocov3} also shows that a Transformer without PEs shows only a small degradation when evaluated with linear probing in a contrastive learning setup. MP3 confirms the hypothesis that much of the Transformer's power lies in its ability to model the co-occurrence of input tokens. In particular, our pretraining method does not use or train PEs at all (instead of randomly shuffling input tokens while using PE), and still performs competitively compared to other baselines.\footnote{Note that for Speech some positions are needed to be added~-- otherwise, the pretraining task is too hard for the model  to solve.}

\textbf{Contrastive Learning}. This is a family of methods for self-supervised learning, where the learning objective is to assign high similarity to augmented views from the same example \cite{cpc,chen2020simple,mocov3,caron2021emerging}. MP3 differs as it does not rely on data augmentation as the source of training signal, which gives it much more flexibility. Besides, MP3 does not enforce clustering of the representation for different positions within an input, which makes it not suitable for linear probing tasks. These differences also suggest a possibility of combining MP3 and contrastive learning to achieve the best of both worlds. There has also been attempts combining contrastive learning with predictive tasks \cite{dangovski2021equivariant}, which suggests possible ways of combining MP3 with contrastive learning in a similar fashion.

\textbf{Position prediction in NLP}. In concurrent works, the idea of position prediction has also been explored in the NLP domain \cite{cui2022pert,bruel2022relative}. These works, combined with MP3, suggest that position prediction is a promising technique across a wide range of problems.

\section{Method}

\subsection{Architecture}
For Vision, our architecture is based on ViTs \cite{vit}. In a nutshell, ViTs divide an image into non-overlapping patches of a given size (e.g., $16\times 16$). 
A linear projection with shared weights to all image patches to obtain a sequence of image tokens is then applied. 
Token vectors are additively combined with their respective positional embeddings to form the input sequence. Standard self-attention layers are then applied to process the input sequence. 

For Speech, our architecture is based on the vanilla Transformer.
The input to the model is a sequence of frames of 40 mel filterbank cepstral coefficients (MFCCs), computed from 30ms of raw waveforms, strided by 10ms between frames, following~\cite{choi2019temporal}. Each frame is transformed by the same linear projection into the dimension of the transformer model (thus, each frame is treated as a 1D patch). A fixed sinusoidal positional embedding~\cite{transformer} is added to these projected representation and the result is fed into an 8 layer Transformer. As with ViTs, we add a learnable ``cls" token frame at the beginning of the model input sequence. Compared to Vision, in Speech we have a patch that is 1D rather than 2D as we ignore the structure in the frequency domain. Later in the text, we refer to a frame as a patch in the context of Speech tasks.

\begin{table*}[!t]
\caption{Overview of datasets and baseline models. The ViT-S and ViT-B architectures are defined according to \cite{deit}.}
\label{tab:datasets}
\vskip 0.05in
\small
\centering
\begin{tabular}{lccccccc}
\toprule
Dataset &  Input size & \#Examples & \#Classes & Model config & Patch size & Patch stride & \#Positions \\
\midrule
CIFAR-100 & $32 \times 32$ & 50K & 100 & ViT-S & $4\times4$ & 4 & 64\\
Tiny ImageNet & $64 \times 64$ & 100K & 200 & ViT-B & $8\times8$ & 8 & 64\\
ImageNet-1K & $224 \times 224$ & 1.3M & 1K & ViT-B & $16\times16$ & 16 &  196\\
Google Speech Commands & 1s & 22246 & 12 & 8 layer Transformer & 30ms & 10ms & 100\\
\bottomrule
\end{tabular}

\vskip -0.1in
\end{table*}

\subsection{Masked Position Prediction Pretraining}
In the pretraining phase, we apply the same patch projection as standard ViTs but remove the positional embeddings from all the patch representations. This results in a set of patch representations. We next randomly select $1 - \eta$ fraction of patches as ``context patches'', where $\eta$ denotes the masking ratio. We then modify the self-attention layers accordingly, where only the context patches take part in the computation of keys and values; queries are computed for all patches. In other words, we perform cross attention from all input patches to the context patches. With $\eta > 0$, the Transformer needs to formulate a good representation of the input given only a subset of the input patches, while ordering all the input patches. This forces the model to reason about the relationship of the context patches and infer masked patches at the same time. As a byproduct, a high masking ratio effectively reduces the computational cost of the Transformer in the pretraininig phase. 

We attach a linear prediction head after the last attention layer, with input and output dimensions being the feature dimension $d$ and number of patches $n$, respectively. The outputs of the linear head are passed to Softmax to form a distribution over patch positions. The position prediction loss is obtained with the cross entropy between the position index and the prediction head's outputs. See Figure~\ref{fig:model} for an illustration, and Appendix \ref{app:algorithm} for a sketch implementation.

\subsection{Supervised Finetuning}
After the unsupervised pretraining step, we finetune the network with labels. Specifically, we remove the position prediction head, and attach a linear classifier head after the ``cls" token, as in standard ViTs. We also apply randomly initialized (learned) positional embeddings (or fixed sinusoidal) to the patch embeddings, also following standard ViTs. Random masking is disabled during this stage and full self-attention is used. The remaining setting of the finetuning step largely resembles that of the supervised training.

\begin{figure*}[t!]
    \centering
    \includegraphics[width=0.85\textwidth]{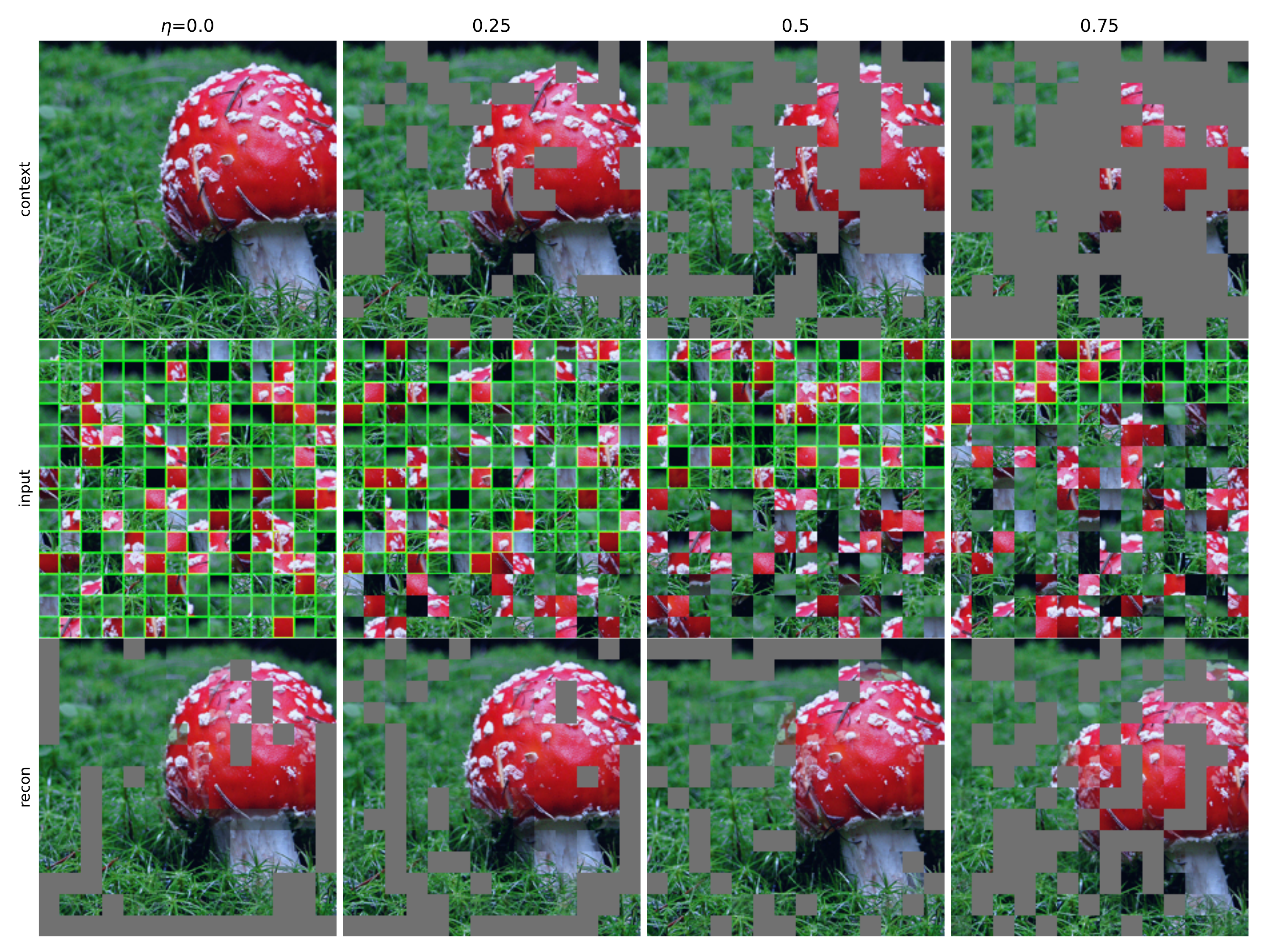}
    \caption{An image from the ImageNet validation set (top left corner) and its reconstructed images for a model trained with $\eta=0.75$. \textbf{Column 1 - 4}: different $\eta$ used at \textit{test time}, ranging in \{0, 0.25, 0.5, 0.75\}. \textbf{Row 1}: the random context patches, placed in their original locations. \textbf{Row 2}: the unordered inputs to the model, with the context patch tokens outlined in green. \textbf{Row 3}: each patch is placed in the predicted position, and patches falling in the same position are averaged. The content in the reconstructed images are still apparent despite distortions. See Appendix \ref{app:reconstruction} for additional examples.}
    \label{fig:recon_vis}
\end{figure*}



\begin{table*}[tb]
\small
    \centering
    \caption{Training time and memory efficiency for MP3, MAE and the ViT-B baseline, while varying the masking ratio $\eta$. MP3 has favorable speed and memory efficiency than both MAE and ViT-B in most settings.}
    \label{tab:efficiency}
    \begin{tabular}{lllll|llll}
         &\multicolumn{4}{c}{Time (Seconds / Iter)} & \multicolumn{4}{c}{Memory (GB / Batch)} \\
         \midrule
         $\eta$ & 0.3 & 0.5 & 0.75 & 0.9 & 0.3 & 0.5 & 0.75 & 0.9\\
         \midrule
         MAE &  OOM & 0.57& 0.47& 0.41 &  OOM & 35.1 & 28.0 & 24.4\\
         MP3 & 0.52& 0.51& 0.46& 0.44 & 30.0& 27.4 & 24.2 & 22.3 \\
         ViT-B & \multicolumn{4}{c}{0.67} &  \multicolumn{4}{c}{33.5} \\
         \bottomrule
    \end{tabular}
\end{table*}

\section{Evaluations}
\label{sec:evaluations}
\subsection{Experimental Setting}
While there has been a lot of interest in scaling Transformers on large datasets in the literature, their performance on small datasets remains under explored. As Transformers tend to overfit easily with pure supervised learning, we believe that it is of great importance to investigate the power of unsupervised pretraining in  scarce data settings. In the domain of vision, we experiment with small to medium sized datasets: CIFAR-100~\cite{krizhevsky2009learning}, Tiny ImageNet \footnote{http://cs231n.stanford.edu/tiny-imagenet-200.zip} and ImageNet-1K~\cite{deng2009imagenet}. In the Speech domain we did not attempt a full blown application of MP3 to Automatic Speech Recognition because the notion of locations is vague, with the streaming nature of the data. Instead we opted here to show proof of concept by applying MP3 to the keyword spotting task, which is a classification problem on a fixed length snippet of audio. We use the Google Speech Commands dataset v1~\cite{google_speech_commands} and implemented our models using the publicly available implementation of TC-ResNet~\cite{choi2019temporal} \footnote{https://github.com/hyperconnect/TC-ResNet}, keeping their audio preprocessing routines, data splits and other details intact. For each dataset above, we choose a baseline Transformer model configuration, the details of which are summarized in Table~\ref{tab:datasets}.




\begin{table*}[t]
\caption{Classification results on CIFAR-100 and Tiny ImageNet. We include a strong ResNeXT baseline \cite{xie2017aggregated,li2021boosting} as a reference in both cases. For baseline ViT-S, we train three versions with absolute, relative and no positional embeddings. We also compare to MOCO V3 \cite{mocov3} and MAE \cite{mae}. MP3 achieves much better results than the supervised learning baseline ViT-S, and is comparable to MOCO V3 and MAE with the same number of pretraining epochs. MP3 without PE achieves surprisingly competitive results in both cases.}
\label{tab:cifar100_tiny_imagenet}
\vskip 0.05in
\small

\centering
\begin{tabular}{lcccc}
\toprule
Method & PT epochs & PE & CIFAR-100 Acc & Tiny ImageNet Acc\\
\midrule
ResNeXT &0 & conv & 82.7 & 72.2\\
\midrule
ViT-S Baseline &0 & absolute & 73.4 & 57.6\\
ViT-S Baseline  & 0& 2D relative & 75.0 & 59.4 \\
ViT-S Baseline &0 & none & 64.6 & 60.0 \\
\midrule
MOCO V3 &2K &2D absolute & 83.3 & 73.4\\
MAE & 2K&2D absolute & 84.5 & \textbf{73.7}\\
\midrule
MP3 & 2K & absolute & 84.0 & 72.8\\
MP3  & 2K & 2D relative & 84.2 & 73.2 \\
MP3 & 2K & none & 82.6 & 68.2\\
\bottomrule
\end{tabular}

\vskip -0.1in
\end{table*}

\begin{table*}[]
    \caption{Classification results on ImageNet-1K. MP3 outperforms the supervised training ViT-B baseline with the same number of total training epochs, which is less overall training cost due to the efficiency of the pretraining phase. Remarkably, MP3 finetuned without any positional information outperforms the full ViT model. MP3's finetuning performance is on par with competitive methods, while with much fewer pretraining epochs.}
    \label{tab:imagenet}
    \vskip 0.05in
    \small
    \centering
    \begin{tabular}{lrrcc}
    \toprule
     Method &  PT Epochs & FT Epochs & PE & Acc\\
     \midrule

     ViT-B \cite{deit} & 0 & 300 & absolute & 81.8\\

     ViT-B \cite{deit} & 0 & 300 & none & 79.1 \\
     ViT-B DINO \cite{caron2021emerging} & 300 & 300 & absolute & 82.8 \\
     ViT-B MOCO V3 \cite{mocov3} & 300 & 150 & 2D absolute & 83.2 \\

     ViT-B BEiT \cite{beit} & 800 & 100 & 2D relative & 83.2 \\

     ViT-B MAE \cite{mae} & 1600 & 100 & 2D absolute & \textbf{83.6} \\ 

     ViT-B MAE \cite{mae} & 150 & 150 & 2D absolute & 82.7 \\
     ViT-B MP3 & 100 & 150 & absolute &83.0 \\
     ViT-B MP3 & \textbf{100} & 300 & none & 81.9 \\
     \midrule

     ViT-L \cite{mae} & 0 & 200 & absolute & 82.6\\
     ViT-L MAE \cite{mae} & 200 & 50 & 2D absolute & 83.3 \\
     ViT-L MAE \cite{mae} & 1600 & 50 & 2D absolute & \textbf{85.1} \\

     ViT-L MP3 & \textbf{150} & 150 & absolute & 83.6 \\
         \bottomrule
    \end{tabular}
\end{table*}


\subsection{Pretraining and Finetuning on Vision Data}
\textbf{Implementation details}. For CIFAR-100, Tiny ImageNet and ImageNet-1k, both our pretraining and finetuning settings largely follow DeiT \cite{deit}, which uses AdamW \cite{adamw} optimizer, weight decay of 0.05, drop path \cite{ghiasi2018dropblock} rate of 0.1, RandAugment \cite{cubuk2020randaugment}, CutMix \cite{yun2019cutmix}, MixUp \cite{zhang2017mixup}, Random Erasing \cite{random_erasing}, Repeated Augmentation \cite{repeated_aug} and label smoothing. In the pretraining phase, we do not use CutMix, MixUP, Random Erasing, Repeated Augmentation and label smoothing. The finetuning phase follows exactly the same protocol as the supervised training recipes suggested in \cite{deit}. We search for optimal $\eta$ for each dataset in the pretraining phase, which is 0.5, 0.8, 0.75 for CIFAR-100, Tiny ImageNet and ImageNet-1K, respectively. The batch size is 256, 512 and 2048, respectively.
 
\textbf{Baselines}. On each dataset, the supervised baseline is trained with strong regularizations. We fix the total training epoch to 400 epochs for CIFAR-100 and Tiny ImageNet, and 300 for ImageNet-1K. We also consider two additional supervised training baselines, one without positional embeddings and another with 2D relative position biases \cite{shaw2018self}. We also consider two Transformer based self-supervised pretraining methods, MOCO V3 \cite{mocov3} and MAE \cite{mae}. In both cases, we use the official code bases and search for the optimal hyper parameter for each case (data augmentation, learning rate for MOCO V3; masking ratio and learning rate for MAE). 

\subsubsection{Pretraining Efficiency}
We first measure the training efficiency of MP3, compared to MAE as well as the supervised training baseline ViT-B. In Table \ref{tab:efficiency} we report the training time (seconds per iteration) and the memory consumption (in gigabytes) on ImageNet-1K with a single A100 GPU. Compared to ViT-B, MP3 has significantly lower time and memory cost across different values of the masking ratio $\eta$. Compared to MAE, MP3 has favorable efficiency for most of the $\eta$ values, espesially when $\eta$ is small.

\subsubsection{Position Prediction}
\begin{figure}[h!]
    \centering
    \includegraphics[width=0.38\textwidth]{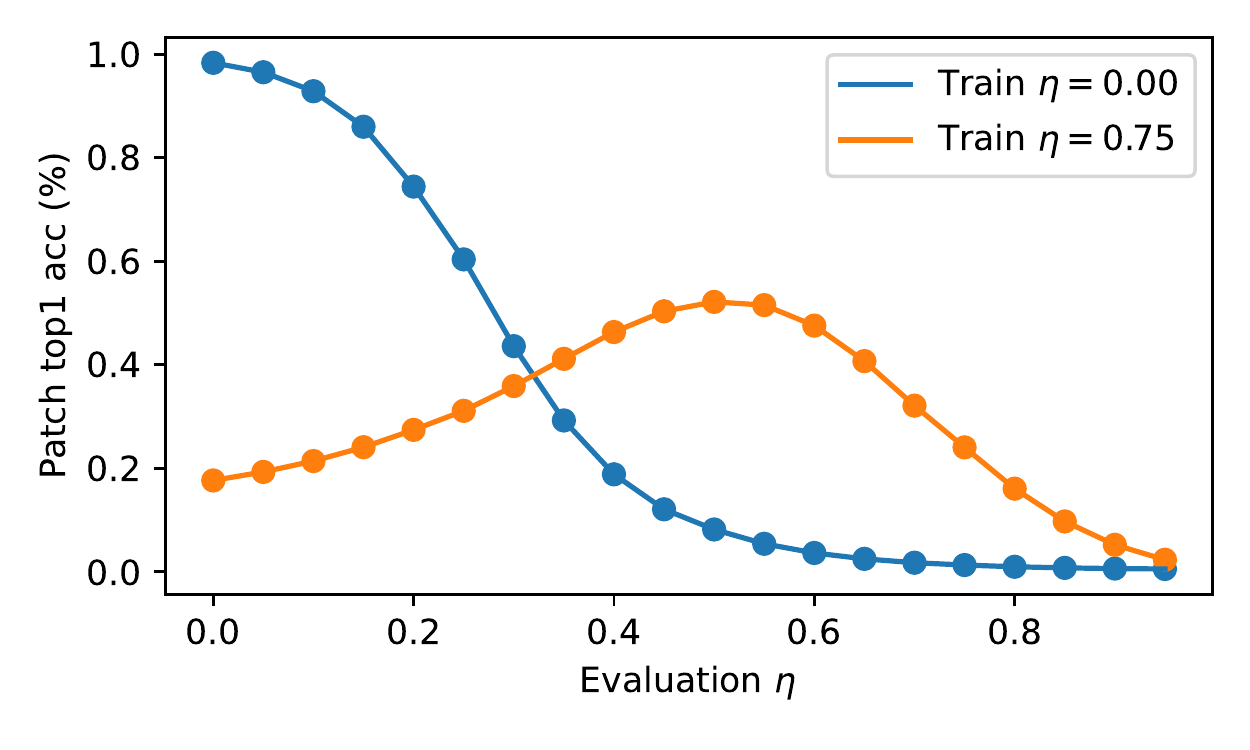}
    \caption{Validation accuracy for \textbf{the position prediction task} on ImageNet-1K for train masking ratios $\eta\in\{0.00, 0.75\}$. The number of total positions is 196. For train $\eta=0$, the position prediction task can be solved near perfectly at evaluation masking ratio $\eta=0$ (which is a standard Jigsaw puzzle), and a large $\eta$ consistently leads to decreasing accuracy. Interesting the converse is true for train $\eta=0.75$, with a patch performance maximum occurring around evaluation $\eta=0.55$.}
    \label{fig:pos_pred_acc}
\end{figure}

Next we examine a Transformer's ability to solve the position prediction task. We show the results for ImageNet-1K where vary the masking ratio in \{0, 0.75\} and train for 100 epochs. We measure the position prediction accuracy on the validation sets with different test $\eta$. The results are shown in Figure \ref{fig:pos_pred_acc}. Interestingly, when trained with $\eta=0$, the Transformers are able to solve the task almost perfectly. Large masking ratio $\eta=0.75$ leads to decreasing accuracy as expected, but the accuracy remains decent up to a high masking ratio. This suggests that there is enough information in the input patches alone to recover their corresponding position information. 

In order to understand the behavior of MP3 with large $\eta$, we show one example in Figure~\ref{fig:recon_vis}. Specifically, we obtain a model trained with $\eta=0.75$, and vary $\eta$ at test time. For each test $\eta$, we generate a random set of context patches, and show the reconstructed images with the predicted positions. We see that the model makes sensible reconstructions, even when the overall accuracy is not high (e.g., with test $\eta=0.75$). This suggests the model can learn to reason effectively about the underlying objects given only a small, positionless subset of input patches. More examples can be seen in Appendix \ref{app:reconstruction}.





\subsubsection{Quantitative Results}
We report the finetuning accuracy in Table~\ref{tab:cifar100_tiny_imagenet} for CIFAR-100 and Table~\ref{tab:imagenet} for ImageNet-1K. In all our experiments, MP3 significantly improves upon the supervised training baseline's accuracy, sometimes by a large margin. Note that we do not change the finetuning hyper parameters, compared to the supervised training baseline, and the gain comes completely from effective pretraining.

Compared to other self-supervised pretraining methods, MP3 achieves comparable results. This is also surprising to some extent, as MP3 does not use or train positional embeddings information in the pretraining phase. We further performed studies of adding zero initialized relative position biases, similar to BEiT \cite{beit}, and not using PE during finetuning. Relative position bias consistently improves upon the absolute PE version, though with a small margin. Interestingly, the version of not using PE shows strong performance, outperforming all the supervised training baselines (including ones with relative position biases). 

Finally, on our largest dataset ImageNet-1K, MP3 requires only 100 pretraining epochs to outperform the supervised trainining baseline, where the total number of epochs are equated. Due to the large masking ratio ($\eta=0.75$) and the use of masked attention, this results in an effective reduction of total training costs (see Table \ref{tab:efficiency} for efficiency measures). We have also experimented with a larger backbone ViT-L. With 150 epochs of pretraining, we are able to outperform the supervise training baseline by 1 point, as well as MAE pretrained with 200 epochs (number taken from the paper).

Note that although MP3 does not outperform the state of the art MAE's performance, we believe that MP3 learns complementary representations. To show this, we performed a simple ensembling test by averaging the outputs of an MP3 and MAE finetuned model from Tab \ref{tab:imagenet} (the ones with 83.0\% and 82.7\% top 1 accuracy, respectively). This results in a strong classifier with 84.0\% accuracy, outperforming MAE pretrained with 1600 epochs. This suggests great potential of potentially combining MP3 and MAE and achieve even greater benefits from pretraining.

\begin{figure*}[t!]
    \centering
    \includegraphics[width=0.49\textwidth]{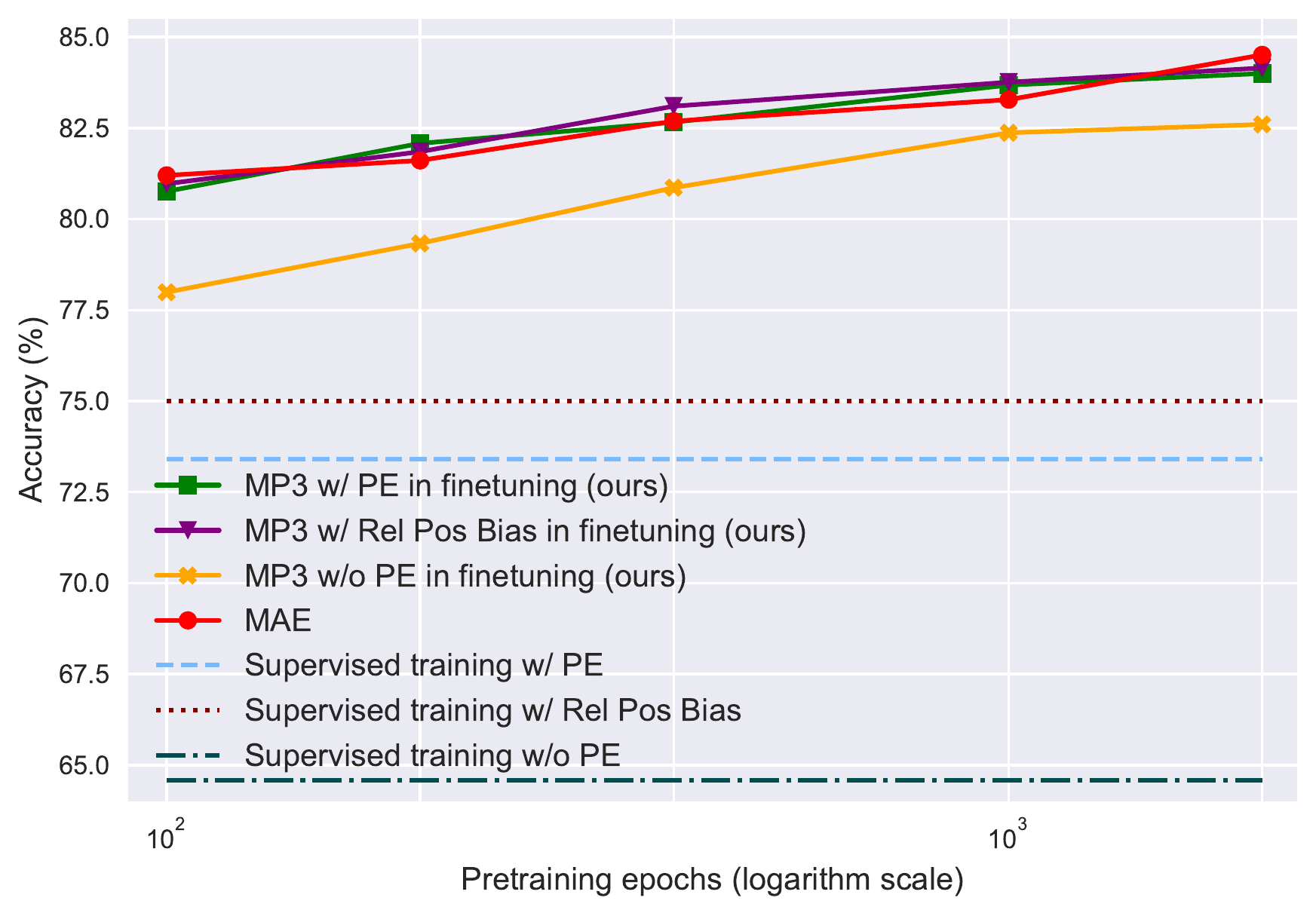}
    \includegraphics[width=0.49\textwidth]{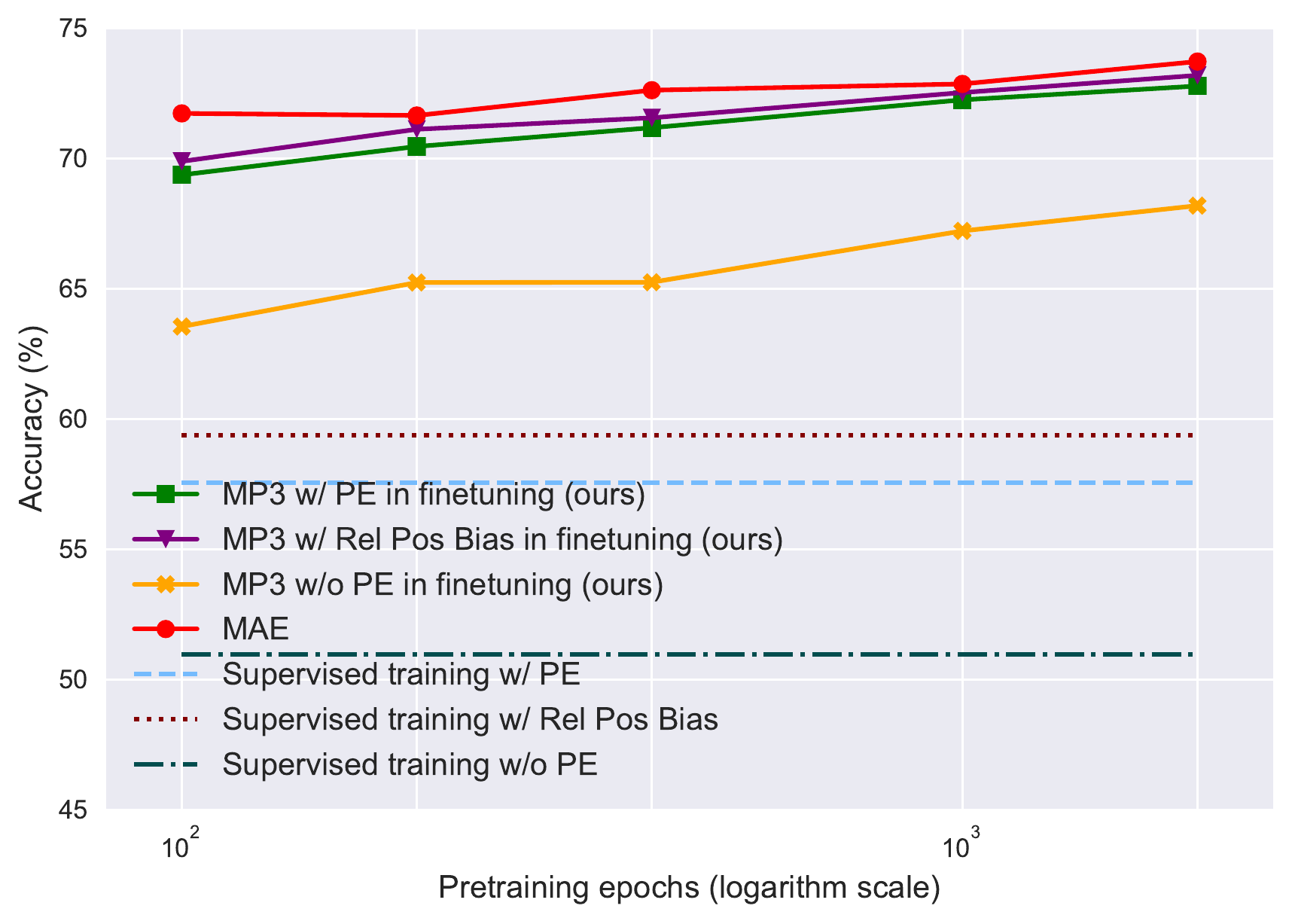}
    \caption{Finetuning accuracy of MP3 on CIFAR-100 (\textbf{left}) and Tiny ImageNet (\textbf{right}) as the pretraining epochs are varied.}
    \label{fig:cifar_tiny_imagenet_results}
\end{figure*}

\subsubsection{Ablations}

\textbf{Pretraining epochs}. We vary the total number of pretraining epochs with everything else fixed, and show the resultant accuracy in Figure~\ref{fig:cifar_tiny_imagenet_results}. We see that MP3 works well with a small number of epochs (e.g., 100) but consistently benefits from more pretraining. 

\begin{figure}
    \centering
    \includegraphics[scale=0.5]{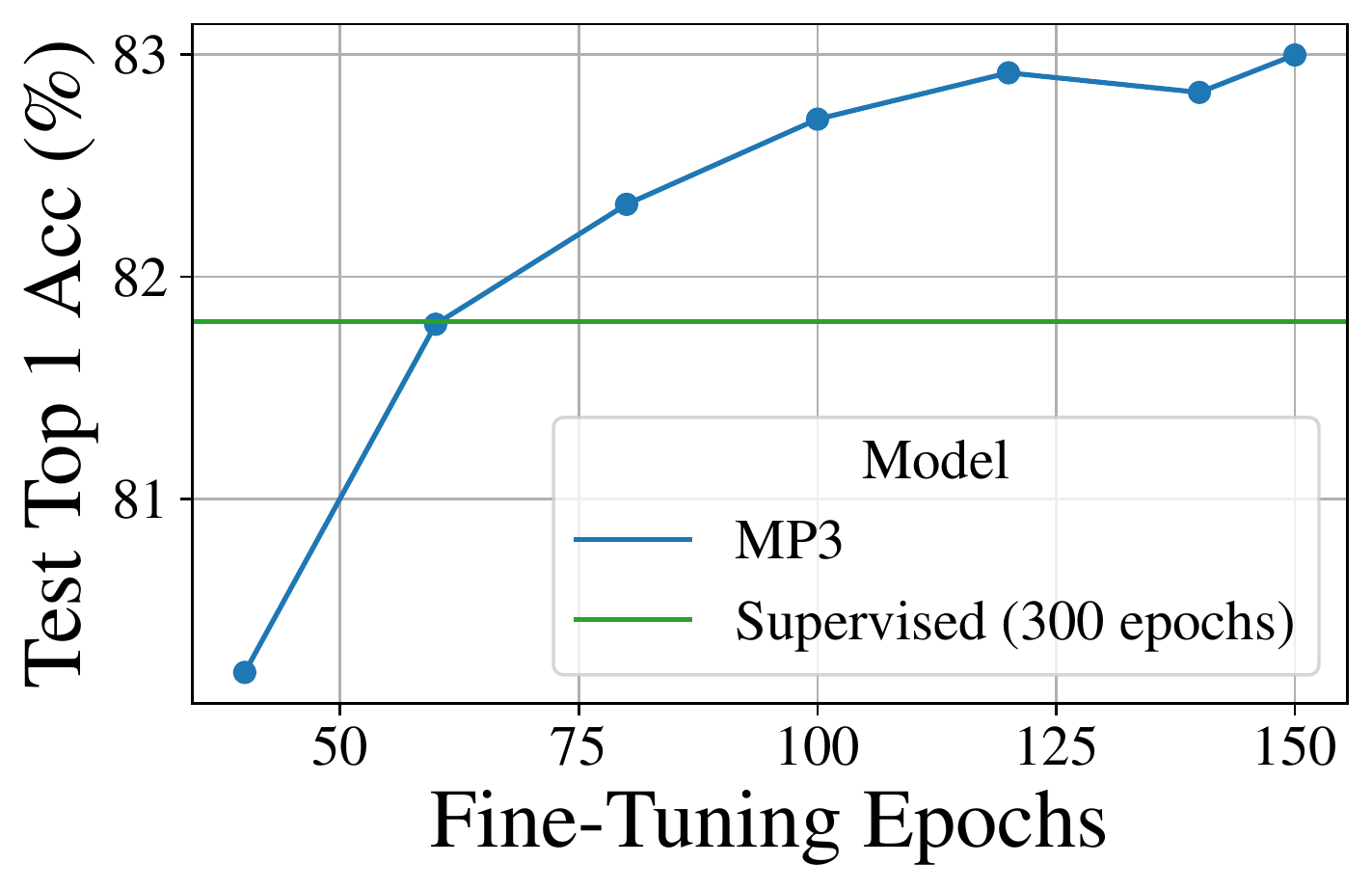}
    \caption{Test top 1 accuracy on ImageNet-1K as the finetunig epochs is varied, with pretraining epochs fixed at 100. MP3 matches the 300 epoch supervised training baseline with as few as 160 total training epochs.}
    \label{fig:ablation_finetune}
\end{figure}

\textbf{Finetuning epochs}. For MP3, the position embeddings are not learned or used during pretraining, which suggests that it can potentially benefit from longer finetuning epochs. To see this, we take a ViT-B based MP3 model pretrained at 100 epochs (see Table \ref{tab:imagenet}) and vary the finetuning epochs. In Figure \ref{fig:ablation_finetune}, we see that this is indeed the case. Moreover, MP3 is able to outperform the supervised training baseline with as few as 60 finetuning epochs (which amounts to 160 total training epochs). This corresponds to an approximate 50\% reduction on the training time. 

\begin{figure}[t!]
    \centering
    \includegraphics[width=0.4\textwidth]{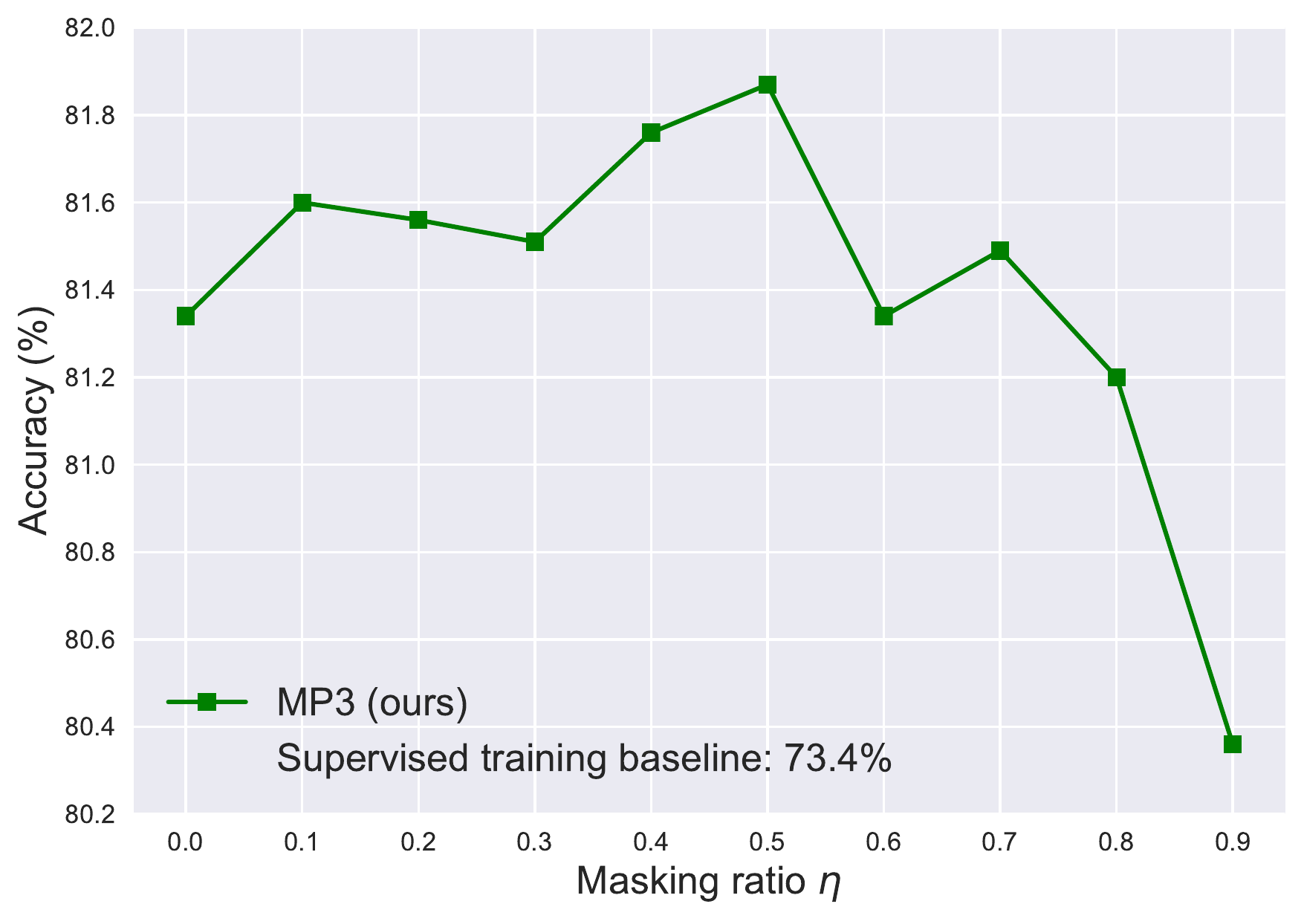}
    \caption{The CIFAR-100 \textbf{finetuning} validation accuracy as the pretraining $\eta$ is varied, with 200 epochs of pretraining. All the settings provide significant improvement to the supervised training baseline, with the performance peaking at 0.5. Extremely large $\eta$ degrades the performance as less contextual information is learned.}
    \label{fig:mask_prob_acc}
\end{figure}

\textbf{Masking ratio}. We evaluate models pretrained with the same number of epochs (200) under different masking ratios. Figure~\ref{fig:mask_prob_acc} shows that there exists an optimal value that induces the highest finetuning accuracy. Extreme large $\eta$ leads to notable degradation, which suggests that it is important to train with a reasonably large context token set. 

\begin{figure}[h!]
    \centering
    \includegraphics[width=0.4\textwidth]{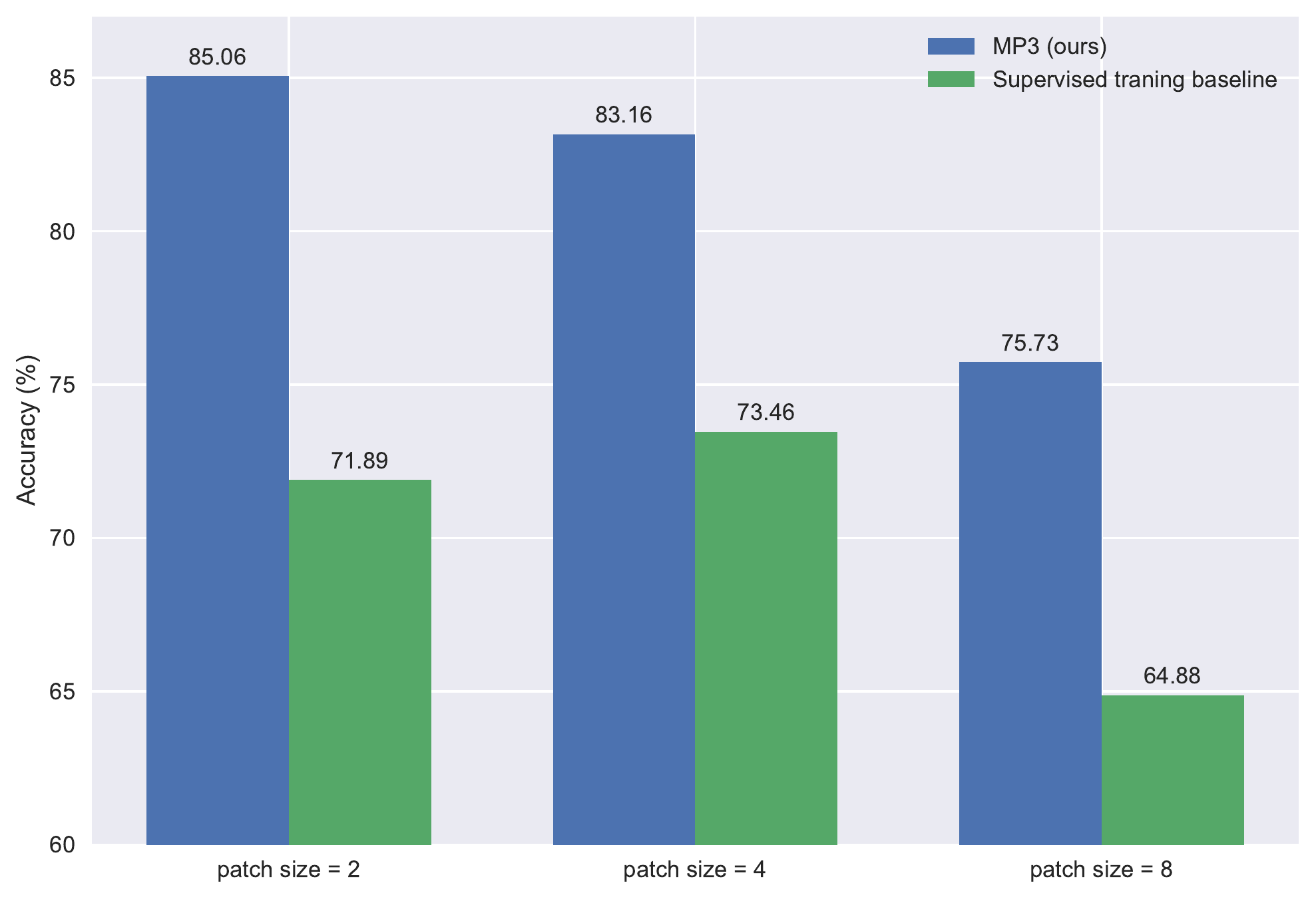}
    \caption{The CIFAR-100 \textbf{finetuning} validation accuracy for MP3 (pretrained for 1000 epochs) and the supervised training baselines, as the patch size is varied. MP3 provides consistent improvements under different patch resolutions.}
    \label{fig:patch_size_acc}
\end{figure}

\textbf{Patch size}. For ViTs, the patch size affects the model's performance. We experimented with two additional patch configurations on CIFAR-100 with the default ViT-S architecture. Figure~\ref{fig:patch_size_acc} shows the accuracy for the supervised training baselines and the finetuning results. We see consistent improvements across small and large patch sizes.

\subsubsection{Visualizing and Understanding Attention}

The improvements demonstrated by MP3 in \cref{sec:evaluations} raise two important questions:
what are the qualitative properties of the attention mechanism that MP3 learns, and which aspects are preserved under finetuning?

\begin{figure*}[ht]
    \centering
    \includegraphics[width=0.9\textwidth]{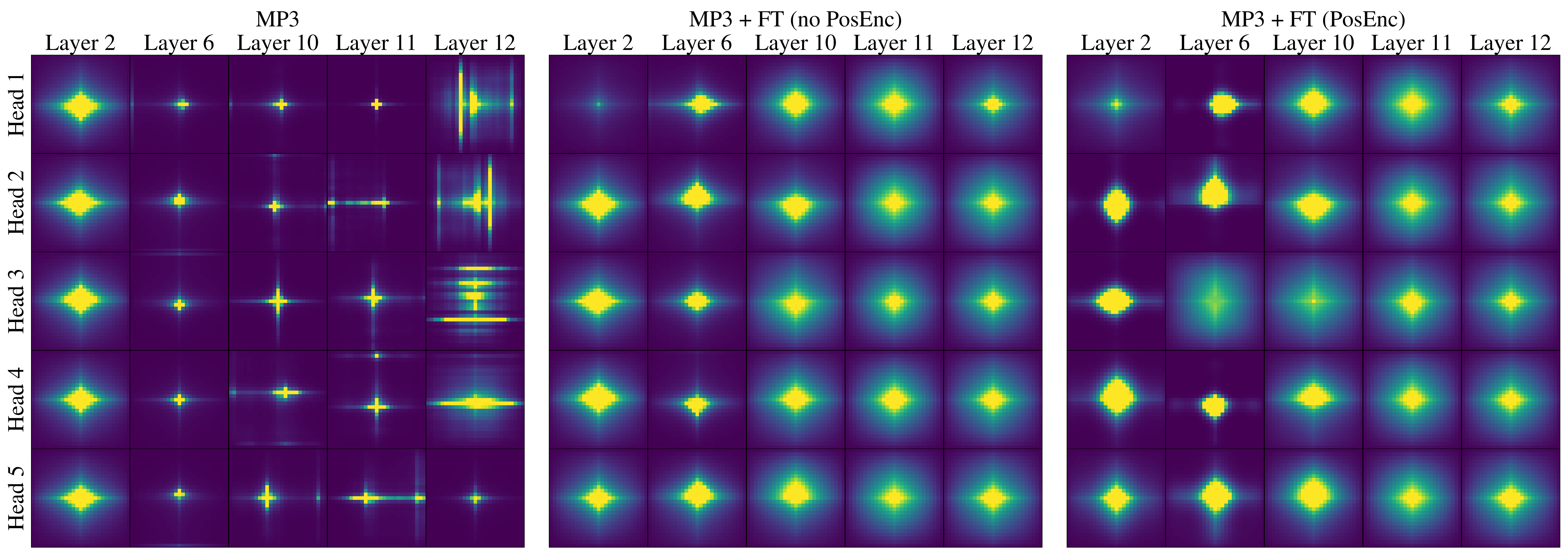}
    \caption{Average relative 2D attention maps for (left) MP3, (center) MP3 + finetuning without positional encoding, and (right) MP3 + finetuning with positional encoding.
    Both fine tuned models are tuned from the same MP3 model (left), which was trained with masking ratio $\eta=0$.
    The heads of the MP3 model are those with the lowest attention entropy 
    $H=-\mathbb E_{p_x} \sum_{i=1}^N\sum_{i=1}^M\alpha^{(x)}_{i,j}\log \alpha^{(x)}_{i,j}/N$,
    and heads of the fine-tuned models are selected to match those of the MP3 model.
    MP3 learns strong localizations in layers 6, 10 and 11, despite not having access to any explicit positional encoding.
    Although localization does occur in early layers of supervised models, we do not see early locality in MP3. 
    We expect this is because of the lack of positional encoding, and a context sufficient for localization has not yet been formed.
    The attention patterns in layer 12 are unlike those of a standard supervised ViT, and we assume they display behaviour specific to the MP3 task.
    Under both finetuning scenario, the later layers are dramatically altered, whereas the earlier layers are less changed. 
    The  primary difference between when using position encoding is that some localization appears in early layers, whereas in it the absense of positional encoding, there is not.
    Each attention map is of size $27 \times 27$, with the class token excluded.
    }
    \label{fig:attention_tuning_effect}
\end{figure*}

We observe that, at all layers, MP3 yields heads that are more local, as well as heads that are more global than those found in supervised ViTs.
Upon finetuning, head locality becomes more similar to that of a supervised ViT, with early layer locality being much less modified than the locality of later layers.
The results for highly local heads at masking ratio $\eta=0$ are illustrated in \Cref{fig:attention_tuning_effect}.
For a full unbiased selection and more details, see \Cref{app:relative-attention}.

\subsection{Pretraining and Finetuning on Speech Data}
For Google Speech Commands we use a Transformer model with 8 self-attention layers, a dropout of 0.1, feature dimension of 32 and fully connected feedforward layer dimension of 64. The model has around 70K parameters in total to be comparable with the smallest convolutional models from~\cite{choi2019temporal}. All pretraining and finetuning models are trained with exactly the same experimental setting as follows. Optimization is done with Adam~\cite{adam} with a batch size of 256 and early stopping is done based on validation accuracy. Warmup of learning is done for 500 updates with a constant learning rate of $10^{-4}$. Subsequently the learning rate is increased to $10^{-3}$ and dropped by a factor of 2 every 10k updates. For supervised baselines and finetuning phase we also use label smoothing (=0.1) for regularization and we train the models for 30K updates.

\begin{table}[t!]
\caption{
Comparison with other baselines on Speech Commands (test accuracy \%).
\label{tab:speech_resuts}}
\begin{center}
\setlength\tabcolsep{5pt} 
\begin{tabular}{ lc } 
\toprule
Model & Accuracy \\
\midrule
TC-ResNet8~\cite{choi2019temporal} & 96.1 \\ 
Transformer & 91.9 \\
Transformer + MP3 & 94.2\\
\bottomrule
\end{tabular}
\end{center}
\end{table}

\begin{table}[t!]
\caption{
Validation accuracy (\%) after finetuning ($\eta=0$) with different amount of pretraining (PT) updates on Speech Commands with 0.05 fraction of the patches being provided positional information.
\label{tab:speech_ablations}}
\begin{center}
\setlength\tabcolsep{5pt} 
\begin{tabular}{ccccc}
\toprule
1K PT & 5K PT & 10K PT & 25K PT & 50K PT \\
\midrule
 91.7   & 93.3 &  94.2 & 93.1 & 93.8  \\
\bottomrule
\end{tabular}
\end{center}

\end{table}

Compared to Vision, the position prediction task (pretraining step) is very hard -- even with $\eta=0$ the top-1 accuracy is only 4\%. Nevertheless, a higher value of top-5 accuracy of 11\% demonstrates that the model is able to learn to roughly position the patches but cannot resolve it further. This result shows the difference between image and audio data: different granularity and locality properties. To simplify the position prediction task, in contrast to vision, we use $\eta=0$ and, moreover, provide positional information for a randomly chosen 5\% of the patches for every sample.  Table~\ref{tab:speech_ablations} shows the results achieved with different amounts of pretraining steps of MP3. It can be seen that 5K steps of pretraining is sufficient to improve model accuracy. The test set result for the base validation model above is 94.2\% which is 2.3\% better than supervised baseline with the same architecture (=91.9\%).

\section{Conclusions}
We have presented MP3, a simple yet effective method for Transformer pretraining. MP3 differs from most other Transformer and token prediction based pretraining method, which bypasses the complexity of designing sophisticated decoder for dense inputs, such as images and speech. MP3 provides competitive performance on small to medium sized data and model sizes, for both Vision and Speech. In particular, MP3 finetuned without position embbedding outperform strong supervised training baselines. We also demonstrate the intriguing properties of the position prediction task, which is of independent interest from the pretraining setting. We believe that the strong performing permutation invariant Transformers will be of great interest to the robust ML community.

There are obvious limitations of this work. First of all, MP3 is not designed to produce linearly separable features which many self-supervised methods excel at (e.g., contrastive learning). Also, despite the high level intuition on sorting input tokens and its relation to semantic understanding, it is not entirely clear how the finetuning benefits from such a pretraining objective. Finally, it is also interesting to test MP3 on NLP applications, and we leave it as future work.


\bibliography{example_paper}
\bibliographystyle{icml2022}

\newpage
\appendix
\onecolumn  

\section{Implementation}
\label{app:algorithm}
Algorithm \ref{alg:mp3} illustrates the MP3 implementation in the pretraining mode. We see that only two simple modifications to the forward pass of a standard Transformer model is needed, which results in a more efficient masked Transformer. The loss is also very easy to compute with the help of a linear prediction head. 

\begin{algorithm}[h]
\caption{Pseudo code of MP3 in a PyTorch-like style, where we ignore the `cls' token for simplicty. In the pretraining phase, we first call \textit{mask\_sample} to randomly sample the context tokens; the context token indices are then passed to \textit{masked\_attention} in each attention block.}
\label{alg:mp3}
\definecolor{codeblue}{rgb}{0.25,0.5,0.5}
\lstset{
  backgroundcolor=\color{white},
  basicstyle=\fontsize{7.2pt}{7.2pt}\ttfamily\selectfont,
  columns=fullflexible,
  breaklines=true,
  captionpos=b,
  commentstyle=\fontsize{7.2pt}{7.2pt}\color{codeblue},
  keywordstyle=\fontsize{7.2pt}{7.2pt},
}
\begin{lstlisting}[language=python]
def mask_sample(x, eta):
    # x: input tokens, shape (batch_size, num_tokens, input_dim)
    # eta: masking ratio in range [0, 1)
    # return kv_ind, indices of context tokens, shape (batch_size, num_context_tokens)
    B, N, D = x.size()
    M = int(N * eta) # number of context tokens
    rand_ind = torch.randn(B, N).argsort(dim=1) # generate a random permutation of positions per input
    kv_ind = rand_ind[torch.arange(B).unsqueeze(1), rand_ind[:, :M]] # get the first M positions per example
    return kv_ind
    
def masked_attention(x, kv_ind):
    # x: input tokens, shape (batch_size, num_tokens, input_dim)
    # kv_ind: indices of context tokens returned by mask_sample, shape (batch_size, num_context_tokens)
    # return y, output of masked attention
    B, N, D = x.size()
    q = query_proj(x) # apply query projection to all tokens
    k = key_proj(x[torch.arange(B).unsqueeze(1), kv_ind])
    v = value_proj(x[torch.arange(B).unsqueeze(1), kv_ind]) # apply key and value projection to context tokens
    y = multi_head_attention(q, k, v) # perform standard multi-head attention
    return y
    
def mp3_loss(x):
    # x: output of the Transformer backbone, shape (batch_size, num_tokens, input_dim)
    # return a scalar loss
    B, N, D = x.size()
    targets = torch.arange(N).repeat(B) # the targets is each patch's original position
    # apply a linear projection to get the predictions
    pred = linear_head(x) # shape (batc_size, num_tokens, num_tokens)
    loss = cross_entropy(pred, targets) # classification across all positions
    return loss

def forward(x, eta):
    # x: input tokens (e.g., image patches), eta: masking ratio.
    # x = x + pos_embed -- we do not use position embeddings
    kv_ind = mask_sample(x, eta)
    x = masked_transformer(x, kv_ind) # with masked_attention
    loss = mp3_loss(x)
    return loss

\end{lstlisting}
\end{algorithm}

\section{Transfer Learning Results}
We further test MP3's ability in Transfer Learning. We obtain a ViT-B model pretrained with 150 epochs with $\eta=0.75$, and finetune it on CIFAR-10 and Stanford Cars \cite{stanford_cars} dataset, which have 50K training examples in 10 classes and 8144 training examples in 196 classes, respectively. We compare with ViT-B, DeiT-B and DINO, all trained with the same architecture. Table~\ref{tab:transfer} shows that MP3 gives competitive performance with supervised and self-supervised models. 

\begin{table}[h]
    \caption{Transfer learning result on CIFAR-10 and Stanford Cars datasets.}
    \label{tab:transfer}
    \vskip 0.05in
    \small
    \centering
    \begin{tabular}{lcccc}
    \toprule
     Dataset &  ViT \cite{vit} & DeiT-B \cite{deit} & DINO \cite{caron2021emerging} & MP3\\
     \midrule
     CIFAR-10 & 98.1 & 99.1 & 99.1 & 98.0\\
     Stanford Cars & - & 92.1 & 93.0 & 91.8\\
   \bottomrule
    \end{tabular}
\end{table}

\section{Layerwise KNN probe}
\label{app:knn}
\begin{figure*}[ht]
    \centering
    \includegraphics[width=0.4\textwidth]{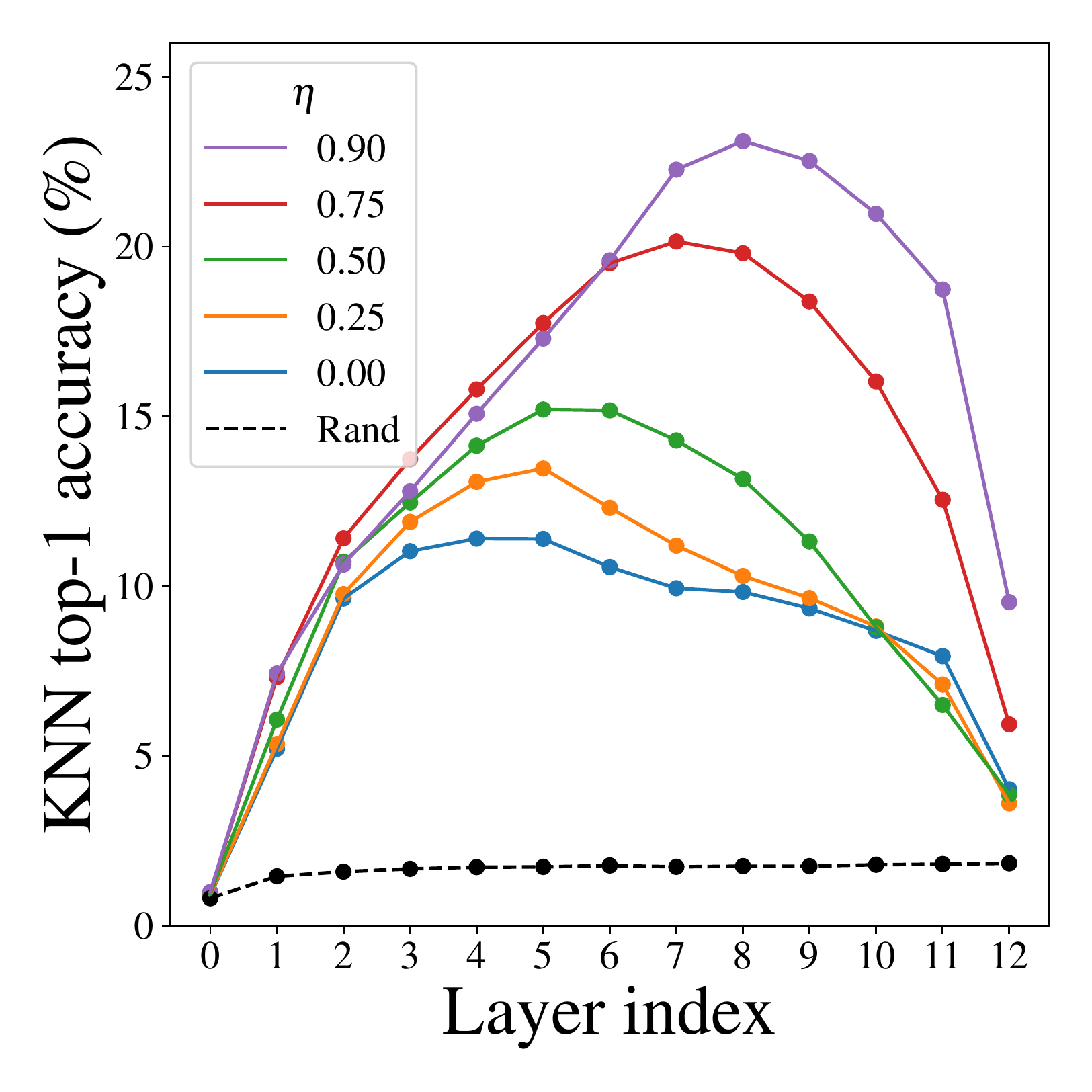}
    \caption{KNN classification accuracy on ImageNet-1K, as the pretraining $\eta$ and the target layer are varied.}
    \label{fig:knn}
\end{figure*}

For self-supervised learning, k-nearest neighbor classification (KNN) is a popular way of testing the linear separability of the pretrained features (which is similar to linear probing). We perform a study on ImageNet-1K, where we vary the masking ratio $\eta$ in pretraining, and examine each layer's average pooled representation with an KNN classifier. The results are shown in Figure \ref{fig:knn}. We see that all trained layers show significantly higher validation accuracy than a random model. There is also a positive correlation between $\eta$ and the peak performance on KNN classification. The optimal layer also appears in the middle, rather than at the very top.

\begin{figure*}[ht]
    \centering
    \includegraphics[width=0.49\textwidth]{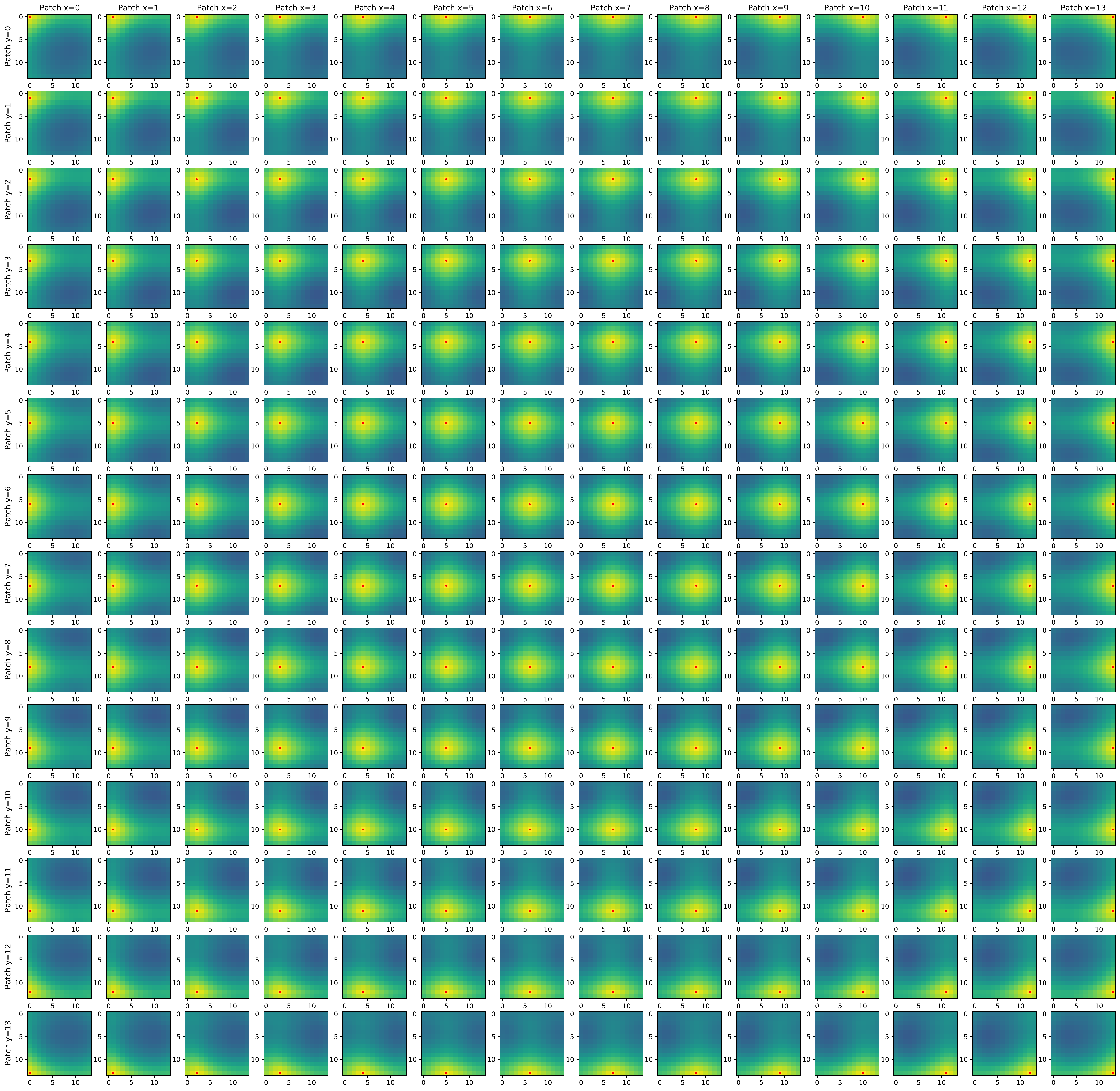}
    \includegraphics[width=0.49\textwidth]{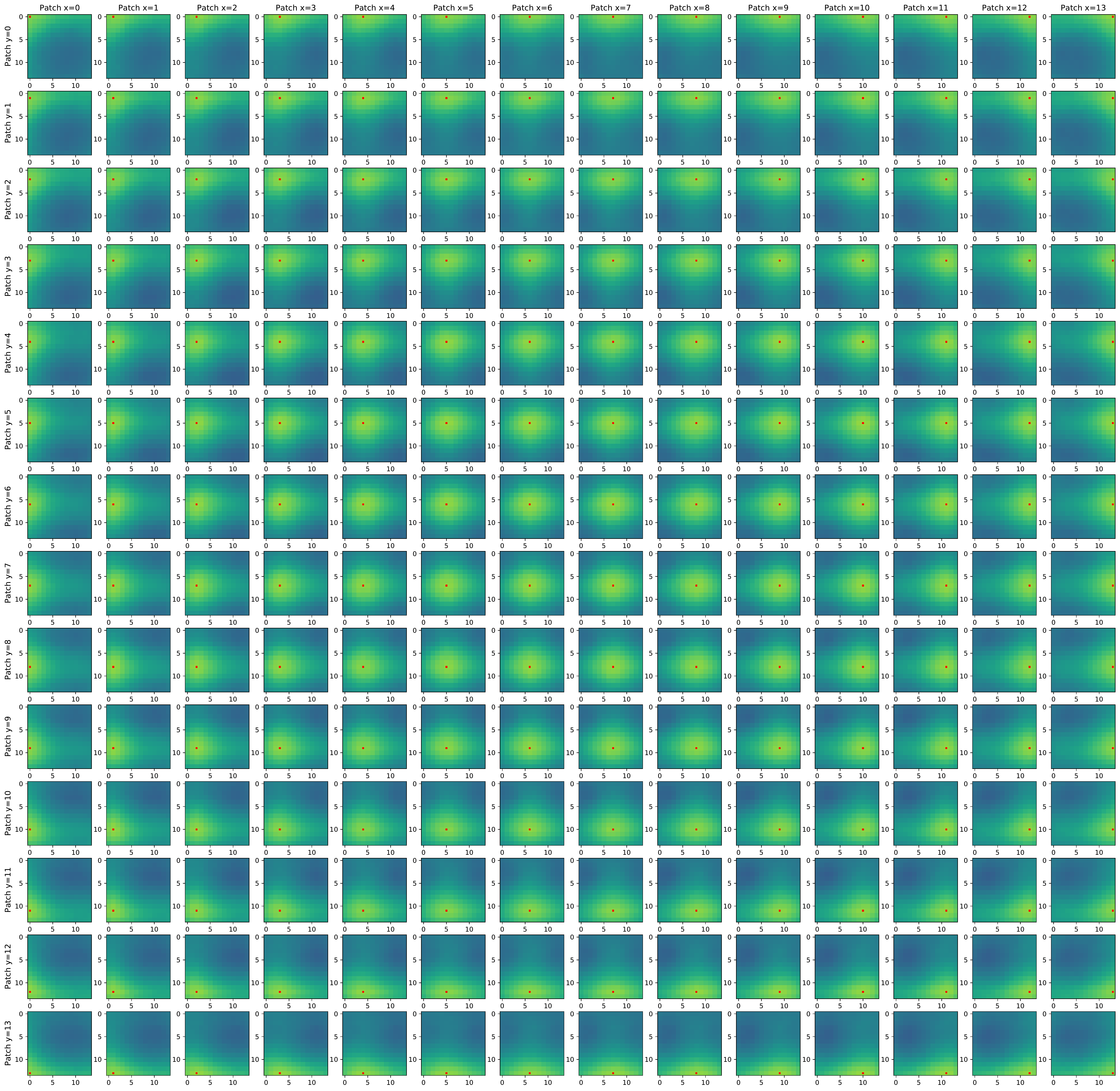}
    \caption{\textbf{Left}: average position correlation within the same image; \textbf{Right}: average position correlation across random image pairs.}
    \label{fig:correlation}
\end{figure*}

\section{Feature Visualization}
We also visualize the correlation pattern of the representations within an image, and across images. To do so, we conduct two experiments. In the first one, we compute the Pearson Correlation of the last layer's representation between each position pair within the same image, averaged across the ImageNet-1K validation set. In the second, we compute the correlation between each position pair of two \textit{random} images. Each experiment results in a correlation matrix of $196 \times 196$, which is reshaped to a $14 \times 14 \times (14 \times 14)$ grid. The results are shown in Figure \ref{fig:correlation}. We see that the representations are biased to their positions. However, there is stronger correlations within the same images than across different ones, which demonstrates that their is an implicit clustering effect of representations within the same image.


\section{Relative attention maps}
\label{app:relative-attention}

\begin{figure}[ht]
    \centering
    \includegraphics[scale=0.12]{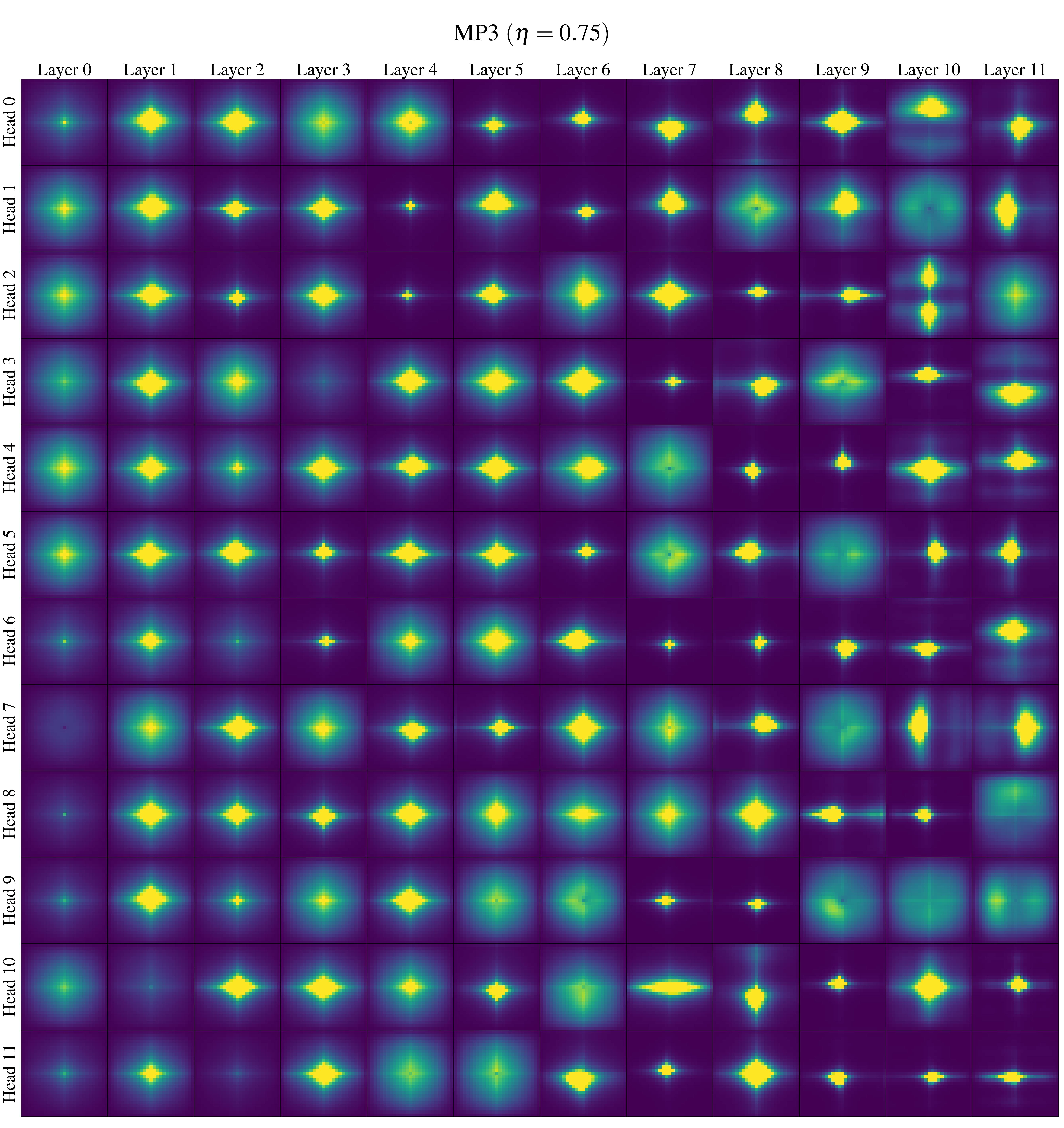}
    \includegraphics[scale=0.12]{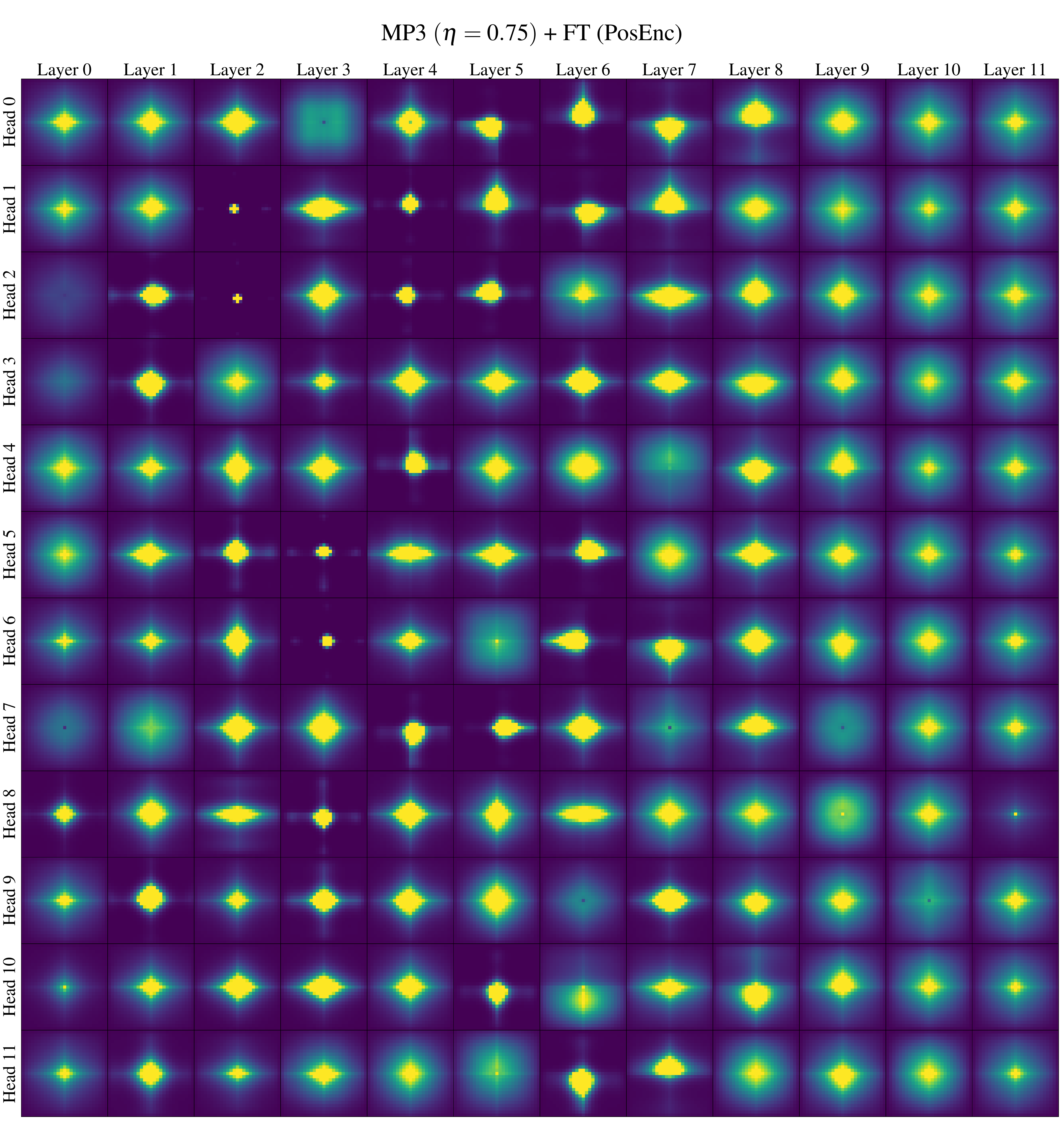}
    \includegraphics[scale=0.12]{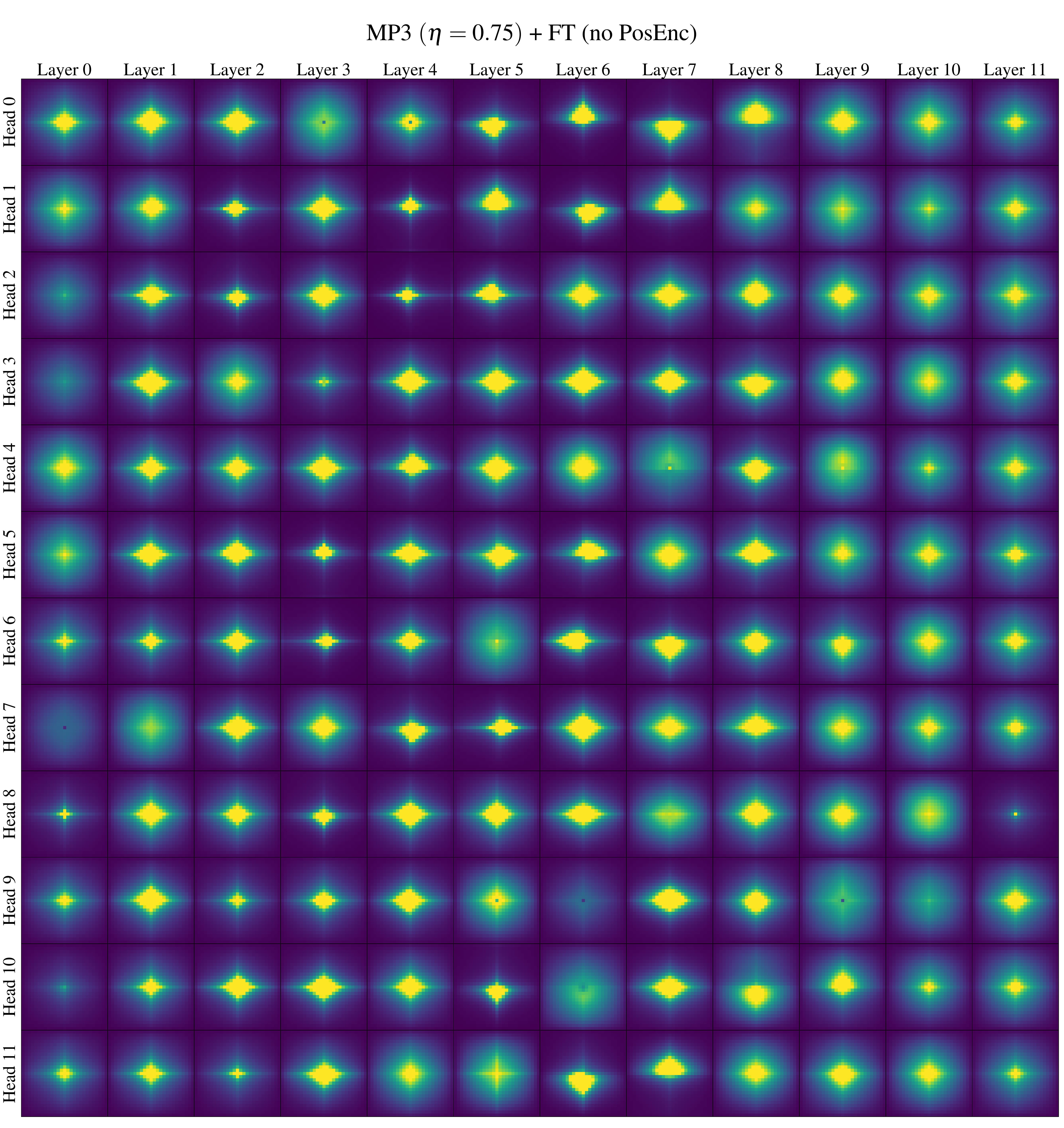}
    \includegraphics[scale=0.12]{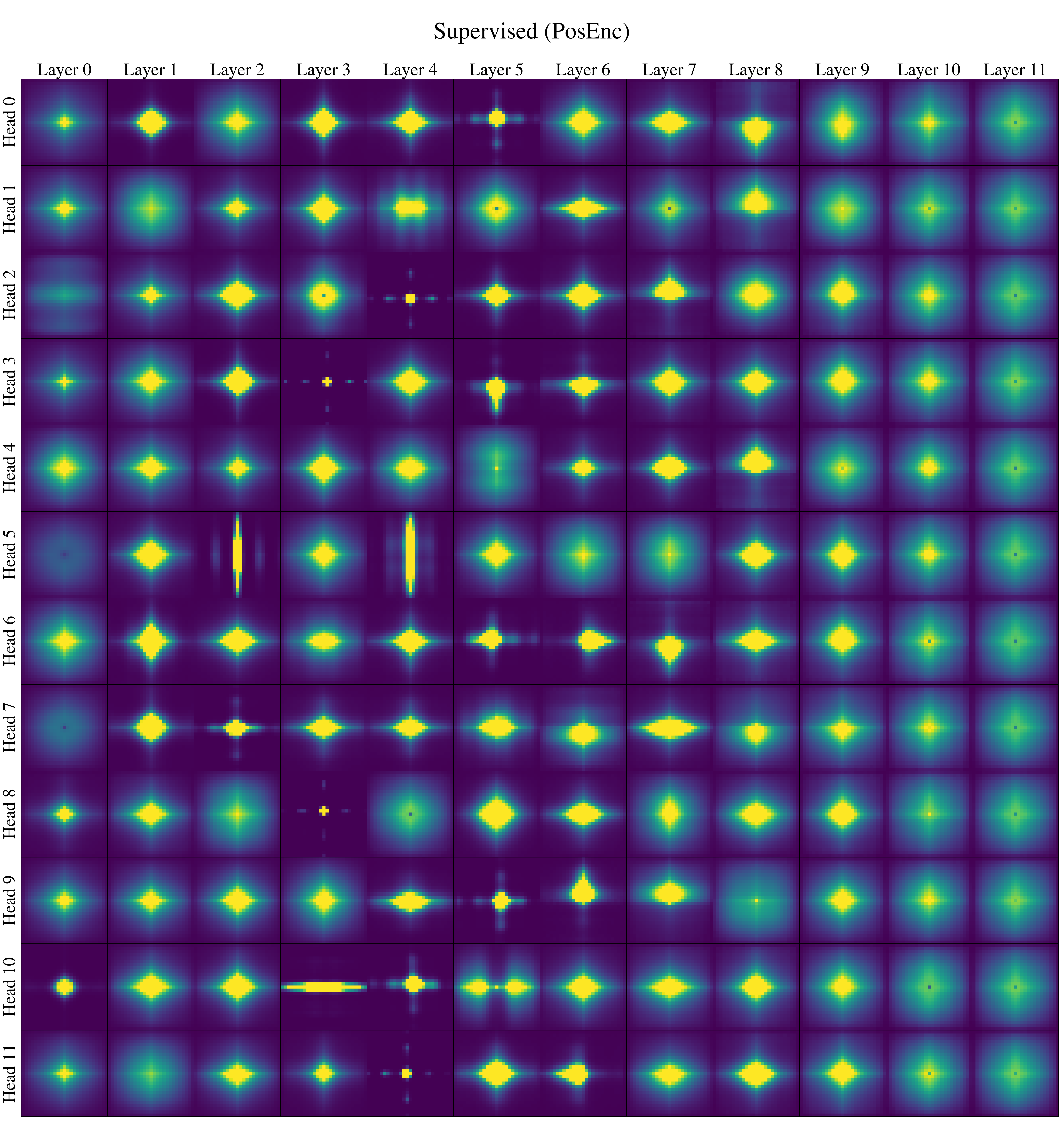}
    \caption{Average relative attention visualization for MP3 pretrained and finetuned models, compared with the supervised training baseline. \textbf{Top left}: MP3 pretrained; \textbf{top right}: MP3 finetuned with PE; \textbf{bottom left}: MP3 finetuned without PE; \textbf{bottom right}: supervised baseline with PE.}
    \label{fig:full_attention}
\end{figure}

Here we present the full (all layers and heads) relative attention maps for MP3 ($\eta=0.75$), finetuning with/without positional encoding, and supervised baseline with positional encoding. The results are shown in Figure \ref{fig:full_attention}. We see that finetuning drastically changes the locality patterns of the last three layers, while lower layers remain similar. 

\section{ImageNet Reconstruction Visualization}
\label{app:reconstruction}

\begin{figure*}[t!]
    \centering
    \includegraphics[width=\textwidth]{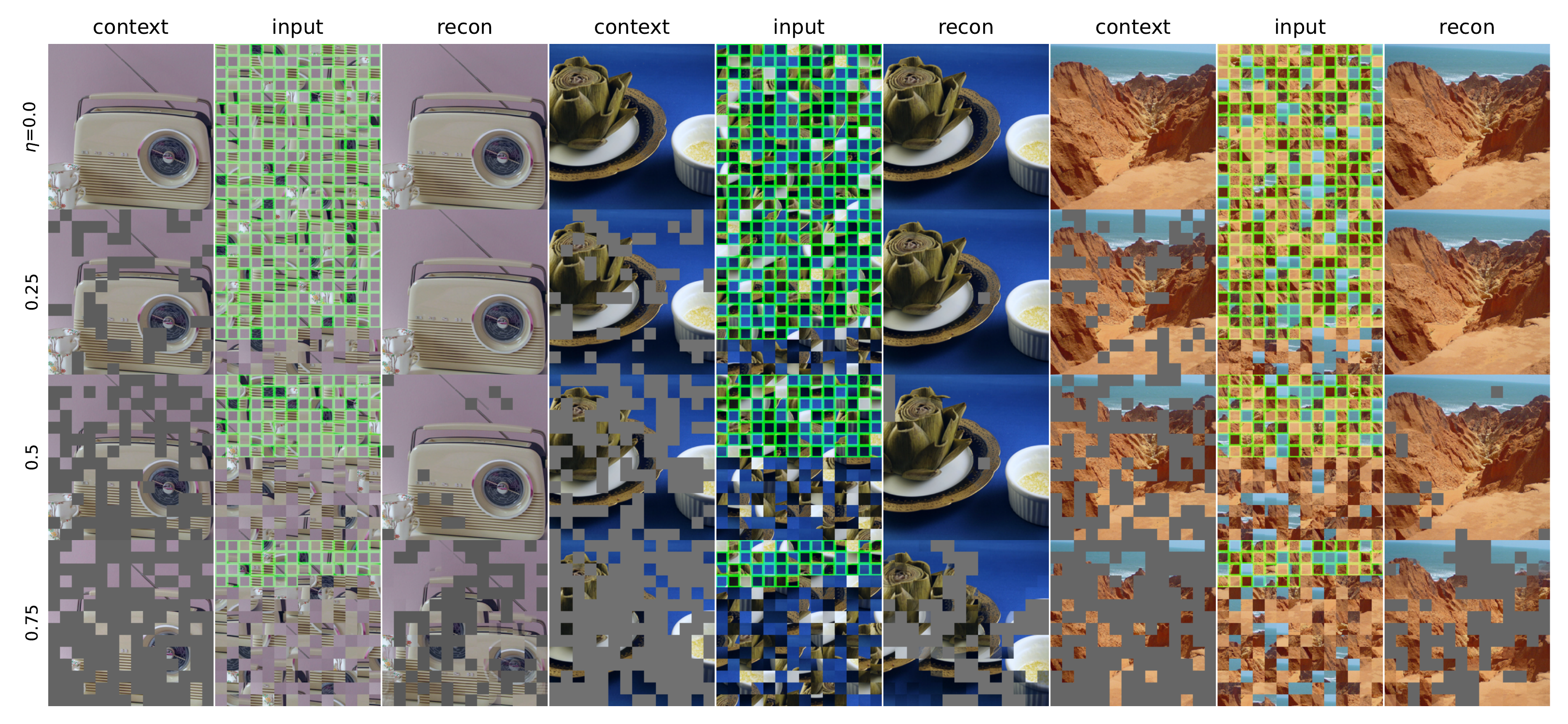}
    \includegraphics[width=\textwidth]{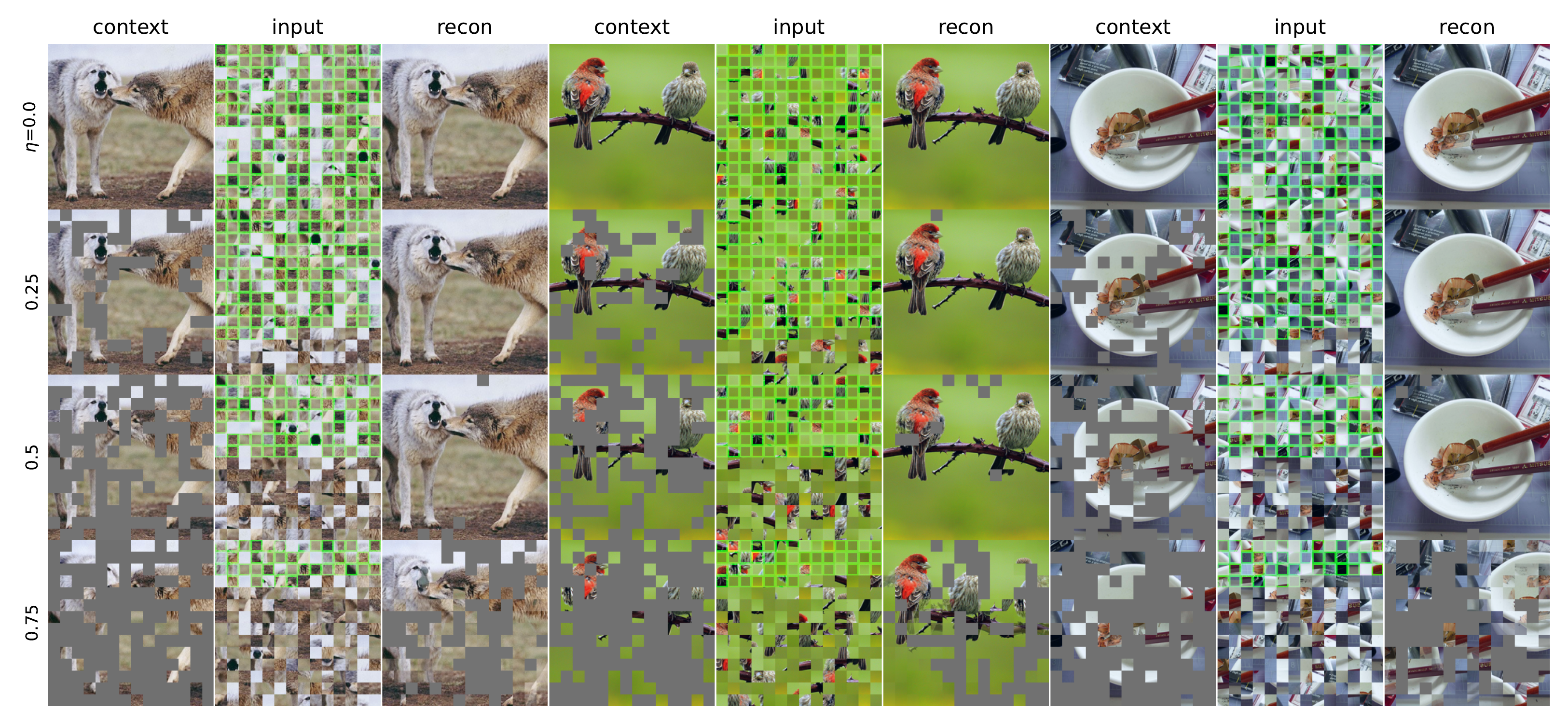}
    \caption{Example reconstructed images from the ImageNet validation set for a model trained with $\eta=0.5$.}
    \label{fig:recon_vis:0.5}
\end{figure*}

\begin{figure*}[t!]
    \centering
    \includegraphics[width=\textwidth]{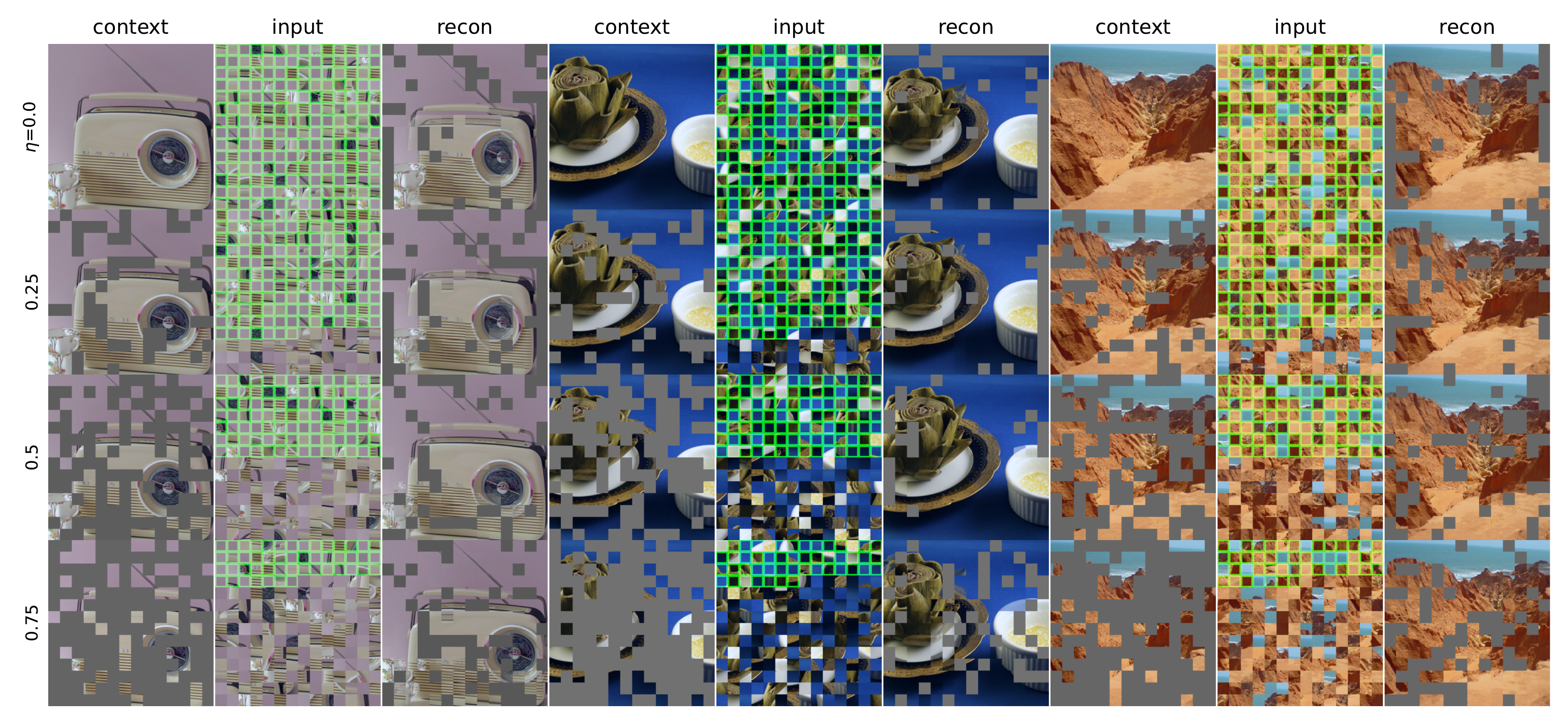}
    \includegraphics[width=\textwidth]{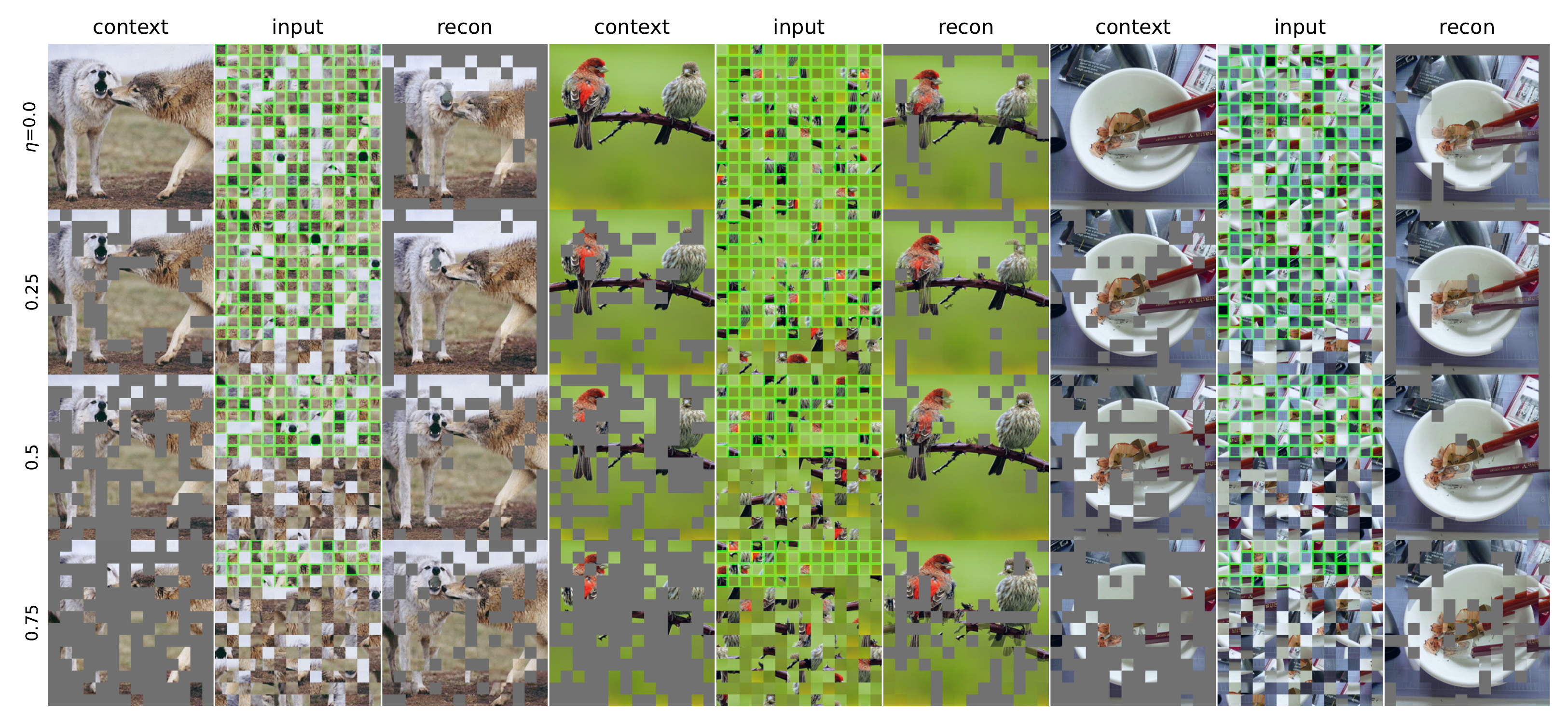}
    \caption{Example reconstructed images from the ImageNet validation set for a model trained with $\eta=0.75$.}
    \label{fig:recon_vis:0.75}
\end{figure*}

As performed in Figure~\ref{fig:recon_vis}, additional images from the ImageNet validation set were used to test the position prediction of a model trained with $\eta=0.5$ (Figure~\ref{fig:recon_vis:0.5}) and $\eta=0.75$ (Figure~\ref{fig:recon_vis:0.75}). Reconstructions were generated by placing each patch in the predicted position, and patches falling in the same position were averaged. Different $\eta$ was used at test time, ranging in \{0, 0.25, 0.5, 0.75\}. In Figure~\ref{fig:recon_vis:0.5} with $\eta=0.5$, the model can accurately predict majority of the patches for $\eta < 0.75$. In Figure~\ref{fig:recon_vis:0.75} with $\eta=0.75$, the patches were not placed in the absolute true location, but they were placed in positions that still made sense semantically.

\begin{figure*}
    \centering
    \includegraphics[width=\textwidth]{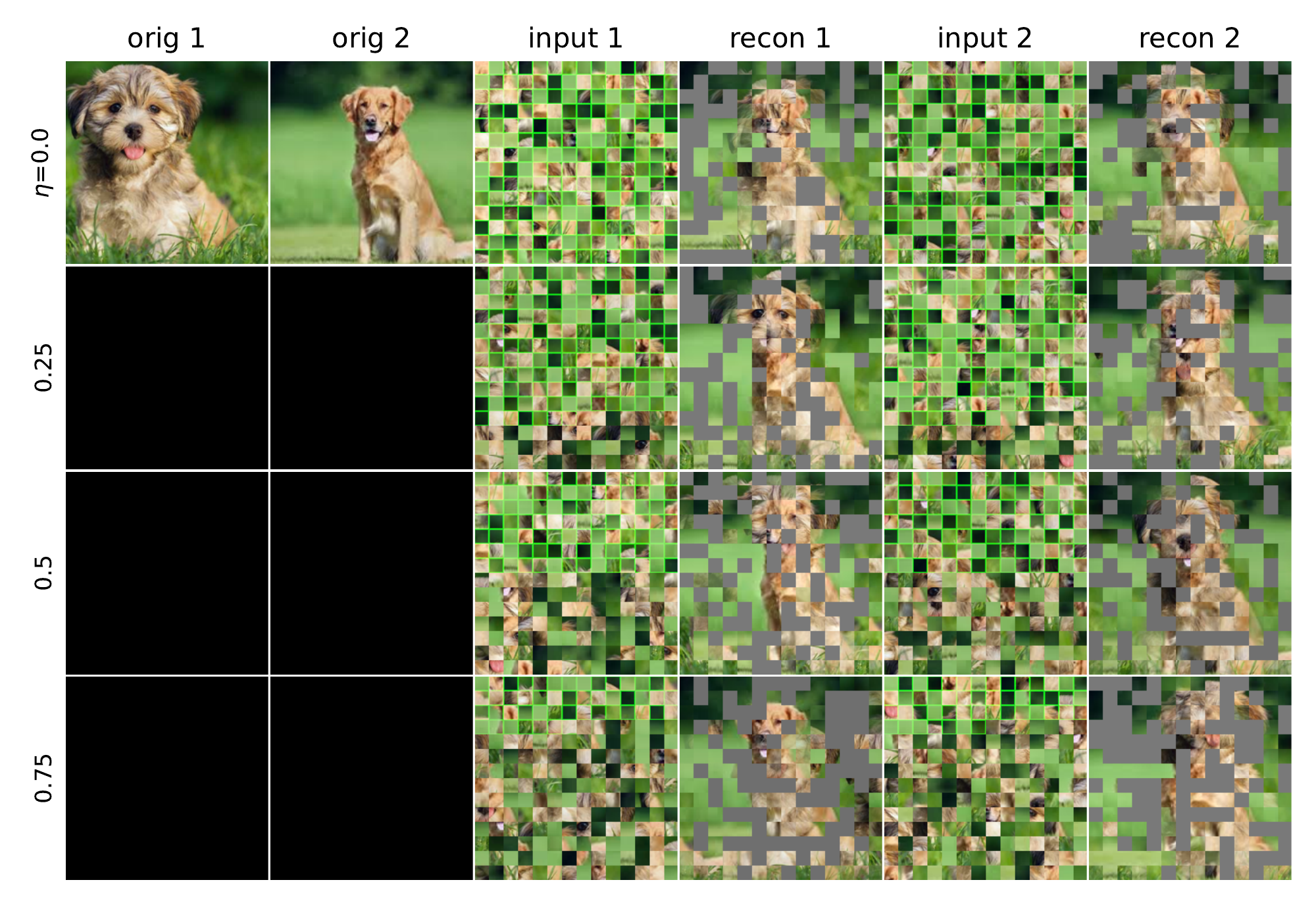}
    \caption{Patches from two images from the ImageNet validation set (top left images) were shuffled together and separated into two distinct sets. \textbf{Rows 1--4}: positions were predicted using the shuffled set of patches with different $\eta$ used at test time, ranging in \{0, 0.25, 0.5, 0.75\}. \textbf{Columns 3 \& 5}: the unordered inputs to the model, with the context patch tokens outlined in green. \textbf{Columns 4 \& 6}: each patch was placed in the predicted position, and patches falling in the same position were averaged. Coherent dog-like animal can be seen in the final reconstructions with the background 
    placed around the dog.}
    \label{fig:recon_vis:shuffle_dog}
\end{figure*}

\begin{figure*}
    \centering
    \includegraphics[width=\textwidth]{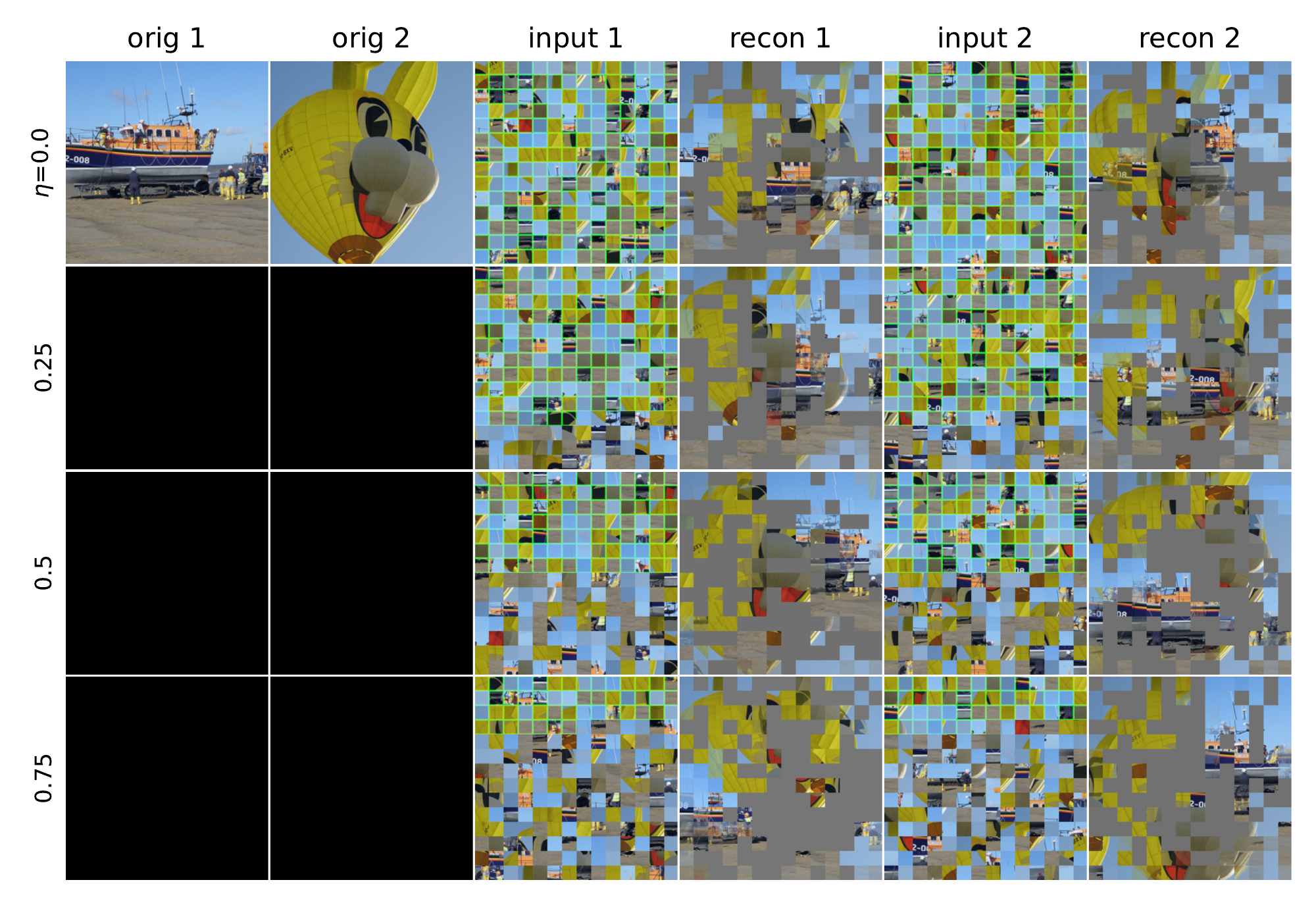}
    \caption{Patches from two images from the ImageNet validation set (top left images) were shuffled together and separated into two distinct sets. \textbf{Rows 1--4}: positions were predicted using the shuffled set of patches with different $\eta$ used at test time, ranging in \{0, 0.25, 0.5, 0.75\}. \textbf{Columns 3 \& 5}: the unordered inputs to the model, with the context patch tokens outlined in green. \textbf{Columns 4 \& 6}: each patch was placed in the predicted position, and patches falling in the same position were averaged. In this example, similar patches were placed closer together. In the last column of Row 3, a coherent boat-like structure was reconstructed in the lower left region of the image.}
    \label{fig:recon_vis:shuffle_boat}
\end{figure*}

To visualize the role of the context patches with the query patches, patches from two different images in the ImageNet validation set were shuffled together and separated into two distinct sets. A random subset of the patches were used as context patches to predict positions for both context and query patches. The final reconstructions are visualized in Figures.~\ref{fig:recon_vis:shuffle_dog} and \ref{fig:recon_vis:shuffle_boat}. In Figure~\ref{fig:recon_vis:shuffle_dog}, the two original images of dogs look visually similar in content, composition, and color distribution. The resulting images created a dog-like animal in the center of the image. In Figure~\ref{fig:recon_vis:shuffle_boat}, two different contents were mixed together: boat in one image and a hot air balloon in another. Similar patches were grouped together creating coherent boat-like structure in one part of the image and a balloon-like structure in another part.

\end{document}